\documentclass{article}
\usepackage[preprint]{log_2024}						

\usepackage{booktabs}						
\usepackage{multirow}						
\usepackage{amsfonts}						
\usepackage{graphicx}						
\usepackage{duckuments}						
\usepackage{amsthm}
\usepackage{algorithm}
\usepackage{algorithmic}
\usepackage{wrapfig}
\usepackage{microtype}      
\usepackage{xcolor}         
\usepackage{amsmath} 

\usepackage{arydshln}
\usepackage{mathrsfs}
\usepackage{colortbl}

\newtheorem{definition}{Definition}[section]
\newtheorem{lemma}{Lemma}[section]

\newtheorem{prop}{Proposition}[section]

\usepackage[numbers,compress,sort]{natbib}	
\newcommand{\mb}[1]{\textcolor{black}{#1}}

\title[Smoothed Graph Contrastive Learning via Seamless Proximity Integration]{Smoothed Graph Contrastive Learning via Seamless Proximity Integration}

\author[M. Behmanesh and M. Ovsjanikov]{%
Maysam Behmanesh \\
LIX, École polytechnique, IP Paris\\
\email{behmanesh@lix.polytechnique.fr}\And
Maks Ovsjanikov \\
LIX, École polytechnique, IP Paris\\
\email{maks@lix.polytechnique.fr}
}

\begin{document}

\maketitle

\begin{abstract}
Graph contrastive learning (GCL) aligns node representations by classifying node pairs into positives and negatives using a selection process that typically relies on establishing correspondences within two augmented graphs. The conventional GCL approaches incorporate negative samples uniformly in the contrastive loss, resulting in the equal treatment of negative nodes, regardless of their proximity to the true positive. In this paper, we present a Smoothed Graph Contrastive Learning model (SGCL), which leverages the geometric structure of augmented graphs to inject proximity information associated with positive/negative pairs in the contrastive loss, thus significantly regularizing the learning process. The proposed SGCL adjusts the penalties associated with node pairs in contrastive loss by incorporating three distinct smoothing techniques that result in proximity-aware positives and negatives. To enhance scalability for large-scale graphs, the proposed framework incorporates a graph batch-generating strategy that partitions the given graphs into multiple subgraphs, facilitating efficient training in separate batches. Through extensive experimentation in the unsupervised setting on various benchmarks, particularly those of large scale, we demonstrate the superiority of our proposed framework against recent baselines. The implementation is available at \url{https://github.com/maysambehmanesh/SGCL}.
\end{abstract}

\section{Introduction}
Graph Neural Networks (GNNs) \cite{pmlr-v70-gilmer17a, kipf2017semisupervised,xu2018how} have developed rapidly by providing powerful frameworks for the analysis of graph-structured data. A significant portion of GNNs primarily focus on (semi-) supervised learning, which requires access to abundant labeled data \cite{veličković2018graph,kipf2017semisupervised,pmlr-v202-behmanesh23a}. However, labeling graphs is challenging because they often represent specialized concepts within domains like biology.

Graph Contrastive Learning (GCL), as a new paradigm of Self-Supervised Learning (SSL) \cite{9462394} in the graph domain, has emerged to address the challenge of learning meaningful representations from graph-structured data \cite{9632431,9764632}. They leverage the principles of self-supervised learning and contrastive loss \cite{pmlr-v97-li19d} to form a simplified representation of graph-structured data without relying on supervised data.

In a typical GCL approach, several graph views are generated through stochastic augmentations of the input graph. Subsequently, representations are learned by comparing congruent representations of each node, as an anchor instance, with its positive/negative samples from other views \cite{veličković2018deep,grace2020vf,icml2020_1971}. More specifically, the GCL approach initially captures the inherent semantics of the graph to identify the positive and negative nodes. Then, the contrastive loss efficiently pulls the representation of the positive nodes or subgraphs closer together in the embedding space while simultaneously pushing negative ones apart.

Conventional GCL methods follow a straightforward principle when distinguishing between positive and negative pairs: pairs of corresponding points in augmented views are considered positive pairs (similar), while all other pairs are regarded as negative pairs (dissimilar) \cite{grace2020vf}. This strategy ensures that for each anchor node in one augmented view, there exists one positive pair, while all remaining nodes in the second augmented view are paired as negatives.

In contrast to the positive pairs, which are reliably associated with nodes having a similar semantic, there is a significant number of negative pairs that have the potential for false negatives.
With this strategy, GCL approaches allocate negative pairs between views uniformly, while we intuitively expect that in contrastive loss, misclassified nodes closer to the positive node should incur a lower penalty compared to those located farther away. However, conventional GCL approaches lack a mechanism to differentiate and appropriately penalize misclassified nodes based on proximity.

One early approach for incorporating proximity information in the conventional GCL method can be computing a dense geodesic distance matrix for the entire graph or using spectral decompositions. However, these approaches can become expensive when applied in the context of contrastive learning.
To tackle this problem, we introduce a \textbf{S}mooth \textbf{G}raph \textbf{C}ontrastive \textbf{L}earning (SGCL) method, which effectively integrates the geometric structure of graph views into a smoothed contrastive loss function. This loss function intuitively incorporates proximity information between nodes in positive and negative pairs through three developed smoothing approaches.

To extend the proposed contrastive loss for large-scale graphs, the GCL framework incorporates a mini-batch strategy. The integration of the mini-batch strategy significantly improves the efficiency of the model in handling large-scale graphs, which is a crucial requirement within the vanilla contrastive loss framework.

Our contributions are summarized as follows:

\begin{itemize}
    \item We introduce three formulations for integrating proximity information into the contrastive learning loss, aimed at improving the assignment of positive and negative pairs.

    \item We devise three novel schemes for a graph contrastive loss function (i.e., SGCL-T, SGCL-B, and SGCL-D) that seamlessly integrates node proximity information, overcoming the uniform negative sampling limitations found in conventional GCL methods.

    \item We extend the model for large-scale graphs by incorporating a mini-batch strategy into the proposed GCL framework, enhancing model efficiency and computational scalability.
    
    \item We perform an analytical study, complemented by extensive empirical evaluations for both node and graph classification on various benchmarks, demonstrating the consistent improvement of SGCL over state-of-the-art GCL methods.
\end{itemize}

A comprehensive and detailed explanation of related work is presented in Appendix \ref{app_related_work}.

\section{Background and motivation}

\subsection{Preliminaries}
\label{sec-Preliminaries}
In the domain of unsupervised graph representation learning, we introduce an undirected graph $\mathcal{G} = (\mathcal{V}, \mathcal{E})$, where $\mathcal{V}$ constitutes the node set $\{{v}_1, {v}_2, ..., {v}_N\}$, and $\mathcal{E}$ denotes the edge set, formally captured as $\mathcal{E} \subseteq \mathcal{V} \times \mathcal{V}$. Within this contextual framework, we establish the definition of two pivotal matrices: the feature matrix $\mathbf{X} \in \mathbb{R}^{N\times F}$, wherein each $\mathbf{x}_{i} \in \mathbb{R}^F$ represents the feature vector associated with a distinct node ${v}_i$; and the binary adjacency matrix $\mathbf{A} \in \{0, 1\}^{N\times N}$.

The objective is to develop a GNN encoder $f_{\theta}(\mathbf{X}, \mathbf{A})$ that takes feature representations and graph structural characteristics of the graph as input and generates reduced-dimensional node embeddings $\mathbf{H} = f_{\theta}(\mathbf{X}, \mathbf{A}) \in \mathbb{R}^{N \times F'}$, where $F' \ll F$. Ultimately, the reduced-dimensional node embeddings prove to be invaluable assets in subsequent tasks, particularly in node classification.

\begin{definition}[Positive and negative set]
In the context of the conventional GCL approach with two graph views $\mathcal{G}^{(i)}$ and $\mathcal{G}^{(j)}$, considering an anchor node $v_t^{(i)}$ in view $i$, the \textit{positive set} consists of embeddings $v_p^{(j)}$ in view $j$ that correspond to the same node as $v_t^{(i)}$. Formally, this is expressed as $\mathcal{P}(v_t^{(i)}) = \{ v_p^{(j)} \}_{p=1}^P$, where $P = 1$ because there is only one $v_p^{(j)}$ that corresponds to the $v_t^{(i)}$. Similarly, the \textit{negative set} for $v_t^{(i)}$ includes all embeddings $v_q^{(j)}$ in view $j$ that do not correspond to the same node as $v_t^{(i)}$, formally expressed as \(\mathcal{Q}(v_t^{(i)}) = \{ v_q^{(j)} \}_{q=1, q \neq t}^{Q}\), where $Q = N - 1$ and $N$ denotes the total number of samples in view $j$.

\end{definition}


\subsection{Uniform negative sampling}

Considering the ground truth, positive/negative pairs demonstrate semantic congruence/incongruence, particularly in relation to shared labels with the anchor. These pairs encompass samples affiliated with either the same class (positive) or different classes (negative). 
Nevertheless, in the absence of labeled information, numerous incongruent nodes are inevitably categorized as false negatives, even when they may share semantic similarities with the anchor node \cite{Xia2021ProGCLRH}.

This misalignment of the negative pairs adversely affects the learning process due to its inadvertent impact on the loss function.
Consider the InfoNCE contrastive loss function \cite{Oord2018} for each anchor node  $v_t^{(i)}$. The objective is to minimize the distance between embeddings of positive pair $\{v_t^{(i)},v_t^{(j)}\}$ and simultaneously maximize the distance between embeddings of negative pairs $\{ v_t^{(i)},v_q^{(j)}\}_{q=1,q\neq t}^{N-1}$:

\begin{equation}
\label{eq_co_loss}
\begin{aligned}
    &\mathcal{L}_{\text{InfoNCE}}(v_t^{(i)},V^{(j)})=-\log{\left(\frac{\exp{(\mathbf{h}_t^{(i)}.\mathbf{h}_t^{(j)}/\tau)}}{\exp{(\mathbf{h}_t^{(i)}.\mathbf{h}_t^{(j)}/\tau)}+\sum\nolimits_{q=1,q\neq t}^{N-1} \exp{(\mathbf{h}_t^{(i)}.\mathbf{h}_q^{(j)}/\tau)}} \right)}
\end{aligned}
\end{equation}

\vspace{6pt}

Misalignment in negative pairs $\{v_t^{(i)},v_k^{(j)}\}$ detrimentally impacts the learning process by introducing errors in the loss computation. The misalignment leads to an undesired increase in the loss, hindering the optimization process. Specifically, the GCL model increases the distance between misaligned negative pairs, and inadvertently separates semantically similar samples, leading to a degradation of overall performance.

Essentially, negative pairs in the contrastive loss function are expected to contribute varying significance based on their proximity to the true positive node. However, in the conventional contrastive learning framework, which lacks information about the proximity of these nodes, all $N-1$ negative pairs are handled uniformly. In other words, the conventional contrastive learning approach treats all misclassified nodes equally regardless of whether the misclassification occurs near the true positive or at a significant distance from it. 

\subsection{Motivation and intuition}
\label{motivation_intuition}


The motivation behind the proposed method is that the loss could seamlessly incorporate graph proximity information into the contrastive learning framework. Namely, in standard contrastive learning, if the network makes an error by declaring a false positive, then this error has an equal penalty regardless of where the false positive is in relation to the true positive. 

A straightforward approach to integrate proximity information into the conventional GCL framework is through the computation of a dense geodesic distance or the utilization of spectral decompositions across the entire graph. However, these strategies incur significant computational costs when applied within the context of contrastive learning.

Our high-level intuition involves a smoothed contrastive learning approach that leverages the inherent geometric information within a graph to assign lower penalties for the negatives that are in close proximity to the ground truth positive. 
As such it promotes predictions that are (similarly to conventional CL) either exactly at the ground truth positive, or (differently from conventional CL) at least in the geodesic vicinity of the positive. By introducing this information, we strongly regularize the learning process, thereby improving the overall accuracy.
In the following, we will demonstrate how leveraging the inherent geometric information within a graph can provide additional insights and enhance the performance of the GCL models.

\subsection{Leveraging the advantages of graph geometry }
\label{extendin_graph_geometry}

In conventional contrastive learning models, the positive pairs between two views are represented by a positive matrix $\Pi_{\text{pos}}^{(i,j)} \in {\{0,1\}}^{N \times N}$, where the diagonal elements are '1' and the off-diagonal elements are '0'. The corresponding negative matrix is defined as $\Pi_{\text{neg}}^{(i,j)} = 1 - \Pi_{\text{pos}}^{(i,j)} \in {\{0,1\}}^{N \times N}$, with '0' on the diagonal and '1' in the off-diagonal positions.

We propose a smoothing strategy that goes beyond simple binary categorization of matrices as positive or negative and applies a form of smoothing to the standard contrastive loss. This strategy allows nodes initially categorized as positive or negative to have values ranging from '0' to '1', indicating their degree of association with positive or negative samples, respectively.

\begin{definition}[Smoothing process]
The smoothing process $\mathcal{S}({\Pi}_{\text{pos}}^{(i,j)},\mathbf{A}^{(i)})$ generate a smooth positive matrix $\tilde{\Pi}_{\text{pos}}^{(i,j)}\in {[0,1]}^{N \times N}$ by iteratively updating binary values of $\Pi_{\text{pos}}^{(i,j)}$ based on the neighboring nodes values in the graph structure $\mathbf{A}^{(i)}$ while preserving the underlying graph structure. The corresponding smoothed negative matrix $\tilde{\Pi}_{\text{neg}}^{(i,j)}$ is then computed as $1-\tilde{\Pi}_{\text{pos}}^{(i,j)}$.
    
\end{definition}

In the following, we introduce three formulations for prompting smoothing, including Taubin smoothing \cite{Taubin}, Bilateral smoothing \cite{710815}, and Diffusion-based smoothing \cite{PALMA20143887}.
Our first objective is to enrich both positive and negative pairs by incorporating neighborhood relationships and capturing the broader context of the nodes. Secondly, we aim to demonstrate how these enriched positive and negative sets can lead to a more effective contrastive loss.

\textbf{Taubin smoothing} $\mathcal{S}_T(\mathbf{V},\mathbf{L};K,\mu,\tau)$ involves iteratively performing two stages of filtering utilized Laplacian matrix $\mathbf{L}\in \mathbb{R}^{N \times N}$ to smooth the binary matrix $\mathbf{V} \in \{0,1\}^{N \times D}$ as follows:

\vspace{-10pt}

\begin{equation}
    \label{eq_Taubin}
    \mathbf{V}^{(k+1)} =(\mathbf{I} + \tau \mathbf{L}) ( (\mathbf{I} + \mu \mathbf{L})\mathbf{V}^{(k)})
\end{equation}

\vspace{-5pt}
This process involves the combined operation of two filters, collaboratively leading to the smoothing of the input signal $\mathbf{V}$. The first filter, the negative Laplacian filter ($\mu < 0$), smooths the input signal, while the second, the positive Laplacian filter $(\mathbf{I} + \tau \mathbf{L})$ ($\tau > 0$), prevents oversmoothing by ensuring $\mu < -\tau$. In our approach, we employ symmetrically normalized graph Laplacian $\mathbf{L}= \mathbf{I}-\mathbf{D}^{-1/2}\mathbf{A}\mathbf{D}^{-1/2}$.

\textbf{Bilateral smoothing} $\mathcal{S}_B(\mathbf{V}, \mathbf{A}; \sigma_{\text{spa}}, \sigma_{\text{int}})$ smooths a binary matrix $\mathbf{V}$ by integrating information from nearby nodes, considering both spatial proximity and intensity similarity. Spatial proximity $d_{\text{spa}}(i,j)$ is measured using shortest path distances, while intensity similarity $d_{\text{int}}(i,j)$ is determined by evaluating the similarity in binary values between two nodes, typically quantified using metrics like the Hamming distance. The bilateral filter weight $w(i,j)$ is then computed by:
\vspace{-3pt}
\begin{equation}
    \label{eq_bi_weight}
    w(i,j) = \exp \left( - \frac{d_{\text{spa}}(i,j)}{2 \sigma_{\text{spa}}^2} - \frac{d_{\text{int}}(i,j)}{2 \sigma_{\text{int}}^2} \right),
\end{equation}

where $\sigma_{\text{spa}}^2$ and $\sigma_{\text{int}}^2$ control the smoothing effects for spatial and intensity components, respectively. The smoothed value for node $v_i$ is computed as a weighted average of its $k$-hope neighbor nodes:
\vspace{-3pt}
\begin{equation}
    \label{eq_bi}
    \tilde{\textbf{v}}_i = \frac{\sum_{j \in \mathcal{N}_k(i)} w(i,j) \textbf{v}_j}{\sum_{j \in \mathcal{N}_k(i)} w(i,j)}.
\end{equation}

\textbf{Diffusion-based smoothing} $\mathcal{S}_D(\mathbf{V},\mathbf{A};K,\eta)$ employs the diffusion equation to propagate information among nodes within a graph, effectively smoothing binary values. The process starts with the original matrix $\mathbf{V}$ as the initial condition, where each binary value serves as the initial "heat" at its respective node. The new value for each node is then iteratively updated based on the diffusion equation and the binary values of its neighbors as $\textbf{v}_i^{(k+1)} = \textbf{v}_i^{(k)}+ \eta \bar{\textbf{v}}_i^{(k)}$, where $\bar{\textbf{v}}_i^{(k)}=\sum_{j\in \mathcal{N}(v_i)} \textbf{v}_j^{(k)}$ is the average value of neighboring nodes, and $\eta$ is the diffusion rate applied to determine how much the binary value diffuses from one node to another.

\begin{figure*}[t]
    \centering
    \includegraphics[width=0.8\textwidth]{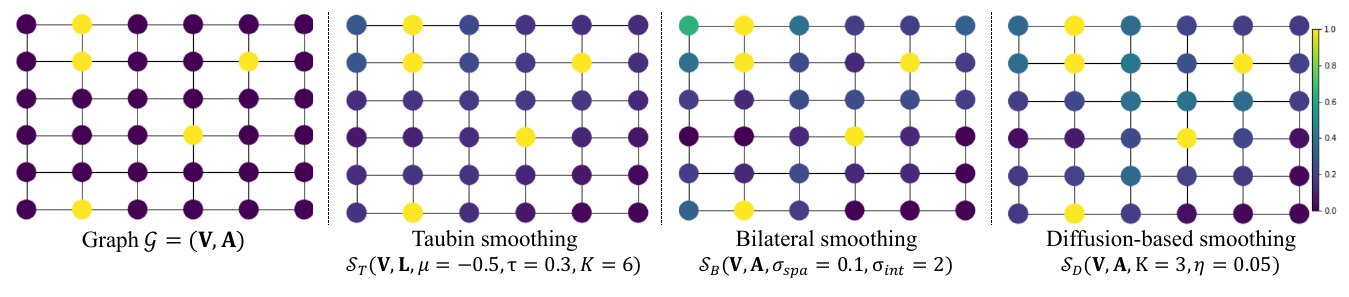}
    \vspace{-10pt}
    \caption{An illustrative example of the efficacy of the smoothing approaches on a grid graph $\mathcal{G}$. We color the grid according to the node value. In the left grid, initial values of 1 are represented in yellow, whereas nodes with zero values are depicted in dark purple. Each smoothing approach modifies the values of the zero nodes according to neighboring information.  }
    \label{fig:img_smooth}
\end{figure*}

\mb{Appendix \ref{ap_algorithms} provides a comprehensive overview of smoothing approaches, including detailed algorithms and a comparative analysis of each method.}  
Figure \ref{fig:img_smooth} illustrates an example of the efficacy of the smoothing approaches. As a simple example, we take a grid graph, and randomly establish a delta function, centered on specific vertices, resulting in the creation of a binary matrix. Subsequently, we employ a variety of smoothing techniques on this binary matrix. Given the uniform neighborhood structure of the grid, the resulting output exhibits a Gaussian-like distribution, which its center aligned to the initial vertex. However, the varied values in the smoothed matrix are indicative of the distinct strategies employed in the smoothing process.

\begin{figure*}[t]
    \centering
    \includegraphics[width=0.7\textwidth]{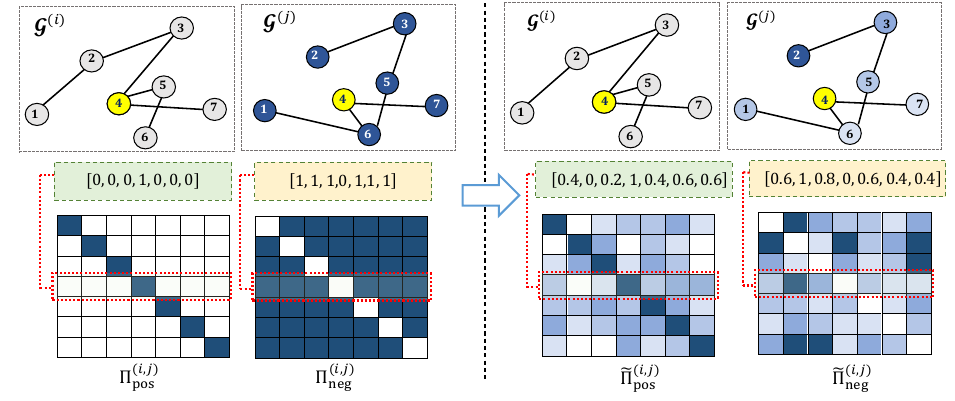}
    \caption{In the general context of conventional contrastive learning approaches, for every anchor node $v_4^{(i)}$ in $\mathcal{G}^{(i)}$, a corresponding positive node $v_4^{(j)}$ exists in $\mathcal{G}^{(j)}$, with all other node pairs being negative (left image). Smoothing techniques, which leverages the  geometry of graph $\mathcal{G}^{(j)}$, effectively extract neighboring node information of node $v_4^{(j)}$ and generate smoothed positive and negative pairs matrices $\tilde{\Pi}_{\text{pos}}^{(i,j)}$ and $\tilde{\Pi}_{\text{neg}}^{(i,j)}$ (right image).} 
    \label{fig:graph_smooth}
\end{figure*}

\subsection{Smoothness promoting in positive and negative sets}

In the context of contrastive learning on graphs, the positive matrix $\tilde{\Pi}_{\text{pos}}^{(i,j)}$ can be considered as a mapping from $\mathcal{G}^{(i)}$ to $\mathcal{G}^{(j)}$, with its rows and columns corresponding to nodes in $\mathcal{G}^{(i)}$ to $\mathcal{G}^{(j)}$, respectively. The goal of the smoothing approach is to extend this mapping to the neighbors of the paired nodes. In this specific context, since the columns of the positive matrix $\Pi_{\text{pos}}^{(i,j)}$ are associated with nodes in $\mathcal{G}^{(j)}$, the smoothing approach utilizes the geometry of graph view $\mathcal{G}^{(j)}$. Similarly, for the positive matrix $\Pi_{\text{pos}}^{(j,i)}$, the smoothing approach utilizes to the geometric properties of the graph view $\mathcal{G}^{(i)}$.

Figure \ref{fig:graph_smooth} illustrates the differences between positive and negative pairs in the conventional graph contrastive learning framework and our proposed smoothed contrastive approach. Notably, when considering a specific anchor node $v_t^{(i)}$ in $\mathcal{G}^{(i)}$ paired with $v_k^{(j)}$ in $\mathcal{G}^{(j)}$, the graph information from $\mathcal{G}^{(j)}$ is employed to generate the smoothed positive and negative pairs matrices $\tilde{\Pi}_{\text{pos}}^{(i,j)}$ and $\tilde{\Pi}_{\text{neg}}^{(i,j)}$.

In the following, we analytically analyze the performance of smooth graph contrastive learning, by defining the following metrics.

\begin{definition}[Dirichlet energy]
The Dirichlet energy of a signal $\mathbf{X} \in \mathbb{R}^{N\times F}$ on the vertices of a graph, defined as $E(\mathbf{X}) = \mathbf{X}^T \mathbf{L} \mathbf{X}=\frac{1}{2} \sum_{i,j} a_{i,j} \| \mathbf{x}_i - \mathbf{x}_j \|^2 $, measures the smoothness of the signal $\mathbf{X}$ over the graph, where $\mathbf{L}=\mathbf{D}-\mathbf{W}$ is the graph Laplacian matrix.

\end{definition}

\mb{A lower Dirichlet energy on a graph indicates that the signal $\mathbf{X}$ varies smoothly with minimal differences between adjacent nodes, aligning well with the graph structure as quantified by the graph Laplacian matrix \cite{6494675}.}

\begin{lemma} [Disparity]
\label{lemma1}
For an encoder \( f_\theta \), the disparity measure of learned features $\mathbf{X} \in \mathbb{R}^{N\times F}$ is defined by the distances of intra-class and inter-class Dirichlet energy as:

\vspace{-10pt}

\[
D_{\text{disparity}}(f_\theta) = \frac{1}{\mid E_{\text{intra}} \mid} \sum_{(i,j) \in E_{\text{intra}}} \Delta_{ij} - \frac{1}{\mid E_{\text{inter}} \mid}\sum_{(i,j) \in E_{\text{inter}}} \Delta_{ij}
\]

where $\Delta_{ij}=\frac{1}{2} a_{ij} \left\| \mathbf{x}_i - \mathbf{x}_j \right\|^2$, and \( E_{\text{inter}} \) and \( E_{\text{intra}} \) denote the sets of edges connecting nodes of different classes and within the same class, respectively. This measure captures the contrast in smoothness between intra-class and inter-class distances in the feature embedding.

\end{lemma}

A lower disparity measure indicates that the encoder produces node representations with greater similarity within the same class and increased distinction between different classes, reflecting more effective self-supervised learning.

\begin{prop}
\label{proposi}
    For two graph encoders $f_\theta$ and $\tilde{f}_{\theta}$ learned using the conventional and smoothed graph contrastive frameworks, respectively, the disparity measure satisfies $D_{\text{disparity}}(\tilde{f}_{\theta}) < D_{\text{disparity}}(f_{\theta})$.
\end{prop}

This proposition indicates that geometry-aware graph contrastive losses enable the learned encoder to more effectively distinguish node representations.



\begin{figure}
    \centering
    \includegraphics[width=0.5\columnwidth]{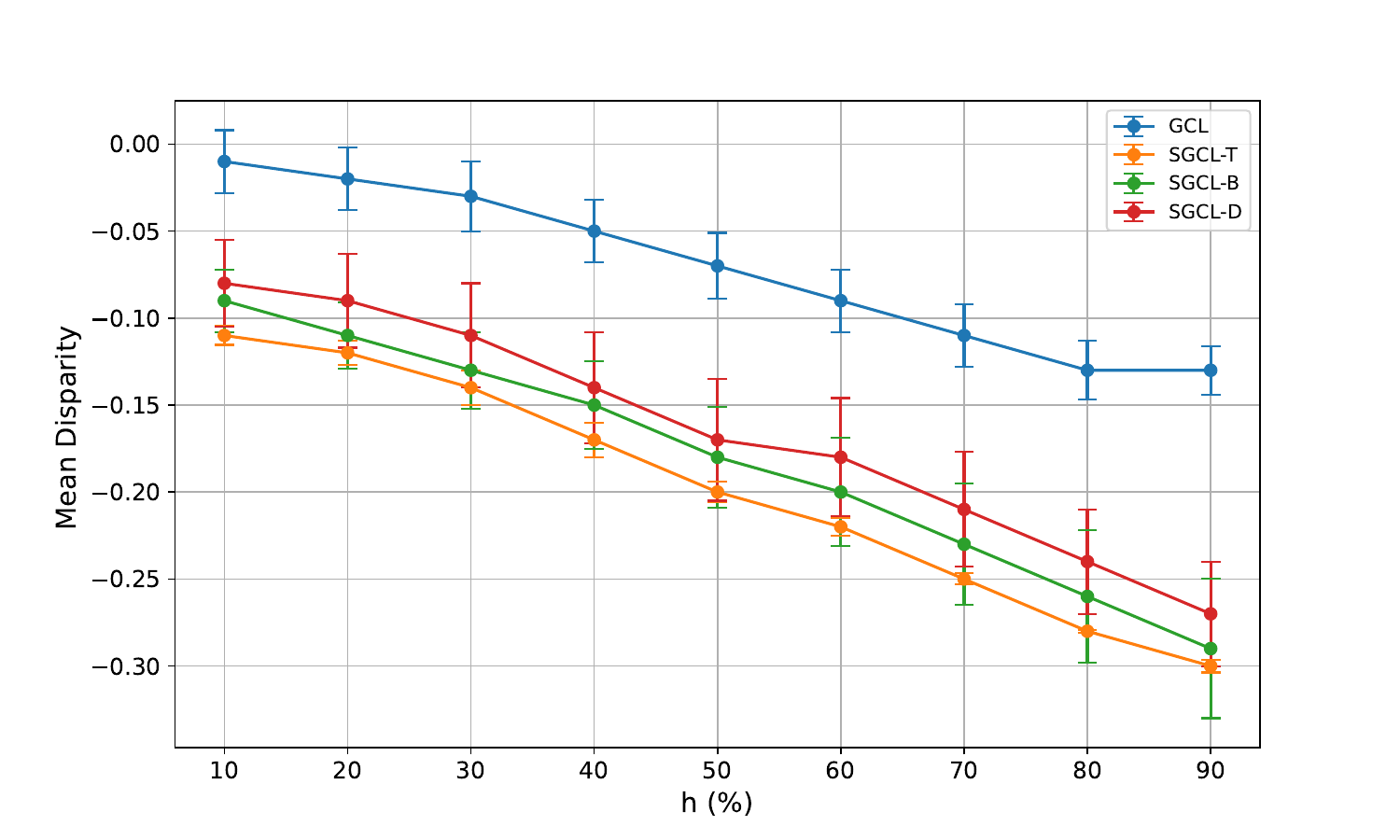}
    \vspace{-10pt}
    \caption{\mb{Comparison of mean disparity among graph encoders, illustrating that all SGCL variants consistently achieve lower values than the conventional GCL approach, which signifies a more effective self-supervised learning framework.}}
    \label{fig:disparity}
\end{figure}

We empirically validate this proposition by computing the disparity measure between two encoders, $f_\theta$ and $\tilde{f}_{\theta}$, used in the proposed SGCL and GRACE frameworks, respectively. Both encoders are applied to the same input graphs across a range of homophily rates. To ensure scale invariance in this comparison, we normalize the feature embeddings. This process removes the influence of the scales in the embeddings, allowing us to focus on the relative differences between embeddings rather than their absolute magnitudes.
We use the graphs from \cite{Karimi2018}, comprising 10 graphs with homophily rates 
$h$ varying from 0 to 0.9. Each graph contains 5000 nodes divided into two classes, sharing the same structure but differing in class labels.
The results in Figure \ref{fig:disparity} indicate that the mean disparity of graph encoders used in all variants of SGCL is consistently lower than the conventional GCL approach reflecting a more effective self-supervised learning framework. 
Additionally, as $h$ increases, disparity measures decrease. This is because of the smoothing strategies that explore positive pairs in the proximity of each anchor node (and similarly for negative pairs). As the homophily rate increases, the number of false negatives inreases, and the role of SGCL in effectively contributing both positive and negative pairs to the contrastive loss becomes more prominent. Further analysis with real-world graphs can be found in Section \ref{emperical_feature_space}.

\section{Method: smoothed graph contrastive learning }

\begin{figure*}
    \centering    \includegraphics[width=0.7\textwidth]{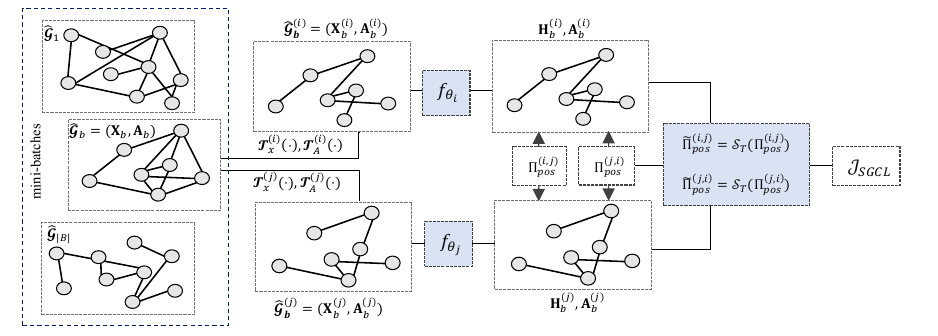}
    \caption{Overview of the Proposed SGCL Model. 
    \mb{The model first generates $|B|$ subgraphs and extracts two distinct views for each subgraph, denoted as $\hat{\mathcal{G}}_b^{(i)}$ and $\hat{\mathcal{G}}_b^{(j)}$. The GCN encoder is then employed to learn feature embeddings $\mathbf{H}_b^{(i)}$ and $\mathbf{H}_b^{(j)}$, respectively. Finally, the smoothed contrastive loss $\mathcal{L}_{\text{SGCL}}$ measures the agreement between these representations by utilizing $\tilde{\Pi}_{\text{pos}}^{(i,j)}$ and $\tilde{\Pi}_{\text{neg}}^{(i,j)}$.}}
    \label{fig:flowchart}
\end{figure*}

We introduce \textbf{S}moothed \textbf{G}raph \textbf{C}ontrastive \textbf{L}earning (SGCL), a novel framework that constructs node embeddings by seamlessly integrating the geometric structure of augmented graphs to ensure a smooth alignment between positive and negative pairs. 
The comprehensive architecture is illustrated in Figure \ref{fig:flowchart}. In the following sections, we outline the processing steps of the proposed framework.

\textbf{Subgraph generating:}
We leverage the random-walk mini-batches generation approach \cite{ZengZSKP20} to generate subgraphs from a given graph. 
More specifically, an entire graph $\mathcal{G}$ is partitioned into a set of $|B|$ mini-batches denoted as $\hat{\mathcal{G}} = \{ \hat{\mathcal{G}}_1, \ldots, \hat{\mathcal{G}}b, \ldots, \hat{\mathcal{G}}_{|B|}\}$, where each $\hat{\mathcal{G}}_b=(\hat{\mathcal{V}_b}, \hat{\mathcal{E}_b})$ represents a sampled subgraph. It is essential to note that the construction of subgraphs varies depending on the specific sampling approach employed.
Leveraging the insights gained from the variance analysis within GraphSAINT \cite{ZengZSKP20}, it introduces a collection of lightweight and efficient mini-batch generation approaches, further detailed in Appendix \ref{sampling_alg}.

\textbf{Generating graph views via augmentation:}
We employ a combination of edge-dropping and node feature masking strategies to generate two distinct graph views for every mini-batch $\hat{\mathcal{G}}_b$, denoted as $\hat{\mathcal{G}}_b^{(1)}$ and $\hat{\mathcal{G}}_b^{(2)}$.
More specifically, in each view $i$, we construct the augmented graph $\hat{\mathcal{G}}_b^{(i)}$ as  $\hat{\mathcal{G}}_b^{(i)}=(\mathcal{T}_x^{(i)}(\mathbf{X}_b), \mathcal{T}_A^{(i)}(\mathbf{A}_b))$, where $\mathcal{T}_x(\mathbf{X})=\mathbf{X}\odot (1- M_X)$ and $\mathcal{T}_A(\mathbf{A})=\mathbf{A}\odot (1-M_A)+ (1-\mathbf{A})\odot M_A$. Here, $M_X \sim \mathcal{N}(0,\Sigma)$ masks original values with Gaussian noise, and $M_A$ utilizes a Bernoulli distribution to randomly drop edges from the adjacency matrix.


\textbf{Encoders:}
The encoder $f_{\theta}$ processes an augmented graph as input, producing reduced-dimensional feature embeddings. We choose the widely adopted Graph Convolutional Network (GCN) \cite{kipf2017semisupervised} as the graph encoder. 
For each view $i$, we employ a dedicated graph encoder $\mathbf{H}=f_{{\theta}_i}(\mathbf{X}, \mathbf{A}): \mathbb{R}^{N\times F} \times \mathbb{R}^{N\times N} \rightarrow \mathbb{R}^{N\times F'}$ that leverages adjacency and feature matrices as two congruent structural perspectives of GCN layers \footnote{For the sake of simplicity, we omit the view index in superscript and the batch index in subscript.}.
The GCN operates across multiple layers, wherein the message-passing process is recurrently applied at each layer. The node representations are updated in a layer-wise manner:
$\mathbf{H}^{(l+1)} = \sigma\left(\tilde{\mathbf{D}}^{-1/2} \tilde{\mathbf{A}} \tilde{\mathbf{D}}^{-1/2} \mathbf{H}^{(l)} \mathbf{W}^{(l)}\right)$, where $\tilde{\mathbf{A}}$ denotes the symmetrically normalized adjacency matrix, calculated as $\tilde{\mathbf{A}}=\mathbf{A} + \mathbf{I}$ with diagonal matrix $\mathbf{I} \in \mathbb{R}^{N\times N} $, $\tilde{\mathbf{D}}_{ii} = \sum_j \tilde{\mathbf{A}}_{ij} \in \mathbb{R}^{N\times N} $ is the degree matrix, $\mathbf{W}^{(l)} \in \mathbb{R}^{F_l\times F_{l+1}}$ is the learned weight matrix for layer $l$, $\sigma$ is activation function, and $\mathbf{\mathbf{H}}^{(l)}\in \mathbb{R}^{N\times F_{l}}$ is the node representation in layer $l$.

\textbf{Smoothed contrastive loss function:}
To end-to-end training of the encoders and promote node representations, we introduce an innovative contrastive loss function. This loss function utilizes a smoothed positive pairs matrix $\tilde{\Pi}_{\text{pos}}^{(i,j)}$ to encourage the agreement between encoded embeddings of two nodes, namely, $v_t^{(i)}$ and $v_p^{(j)}$, in two different views with degree $\hat{\pi}_{\text{pos}}^{(i,j)}(t,p)$, while also distinguish their embeddings with a degree of $\hat{\pi}_{\text{neg}}^{(i,j)}(t,p)= 1-\hat{\pi}_{\text{pos}}^{(i,j)}(t,p)$. The loss function is defined as:
\vspace{-7pt}
\begin{equation}
\label{equ_cont}
\begin{aligned}
   \mathcal{L}_{\text{SGCL}}^{(i,j)} = 
   \parallel {\tilde{\Pi}_{\text{pos}}^{(i,j)}\odot (\mathbf{1}-\mathbf{C}^{(i,j)})} \parallel_F^2 +  \lambda \parallel {\tilde{\Pi}_{\text{neg}}^{(i,j)} \odot \mathbf{C}^{(i,j)}} \parallel_F^2
\end{aligned}
\end{equation}

where $\mathbf{1}$ is a matrix of the same size as $\mathbf{C}^{(i,j)}$ with all elements set to 1 and $\mathbf{C}^{(i,j)}$ is the normalized cosine similarity matrix between the normalized embeddings $\mathbf{\hat{H}}^{(i)}$ and $\mathbf{\hat{H}}^{(j)}$ of identical networks:

\vspace{-9pt}
\begin{equation}
\label{equ_cross}
\mathbf{C}^{(i,j)}=\frac{1}{2}\left(\frac{\mathbf{\hat{H}}^{(i)} \mathbf{\hat{H}}^{{(j)}^T}}{\parallel \mathbf{\hat{H}}^{(i)} \parallel \parallel \mathbf{\hat{H}}^{(j)} \parallel} +1 \right)
\end{equation}

Our objective is to maximize $\mathbf{C}^{(i,j)}$ for positive pairs and minimize $\mathbf{C}^{(i,j)}$ for negative pairs. This is equivalent to simultaneously minimizing $1-\mathbf{C}^{(i,j)}$ for positive pairs and $\mathbf{C}^{(i,j)}$ for negative pairs.

In the proposed contrastive loss function, the first term enforces the stability of the preservation in the embeddings of positive pairs by minimizing the discrepancy between 1 and each element of $\mathbf{C}^{(i,j)}$. This alignment is achieved with the values in the smoothed positive pairs matrix $\tilde{\Pi}_{\text{pos}}^{(i,j)}$, effectively equivalent to maximizing $\mathbf{C}^{(i,j)}$ for positive pairs. Conversely, the second term actively promotes a substantial diversity in the embeddings of negative pairs by minimizing each element of $\mathbf{C}^{(i,j)}$ concerning the values in the smoothed negative pairs matrix $\tilde{\Pi}_{\text{neg}}^{(i,j)}$.

At each training epoch, the smoothed positive pairs matrix $\tilde{\Pi}_{\text{pos}}^{(i,j)}$ is computed by applying one of the smoothing approaches outlined in Section \ref{extendin_graph_geometry}. For example, we apply Taubin smoothing, resulting in the smoothed positive pairs matrix $\tilde{\Pi}_{\text{pos}}^{(i,j)} = \mathcal{S}_T(\Pi_{\text{pos}}^{(i,j)}, \mathbf{L}^{(i)}; \mu, \tau)$, and the corresponding smoothed negative pairs matrix $\tilde{\Pi}_{\text{neg}}^{(i,j)} = 1 - \tilde{\Pi}_{\text{pos}}^{(i,j)}$.
Ultimately, we learn the model parameters by considering all $|B|$ batches within the given graph concerning the overall innovated contrastive loss $\mathcal{J}_{\text{SGCL}}=\frac{1}{2|B|}\sum_{b=1}^{|B|} (\mathcal{L}_{\text{SGCL}}^{(i,j)}+\mathcal{L}_{\text{SGCL}}^{(j,i)})$.

In Equation \eqref{equ_cont}, the hyperparameter $\lambda > 0$ determines the trade-off between two terms during optimization. 
\mb{A comprehensive ablation study on the hyperparameters can be found in Appendix \ref{abblation}.}

\vspace{-10pt}
\section{Experiments}
\vspace{-5pt}
We conduct empirical evaluations of our proposed SGCL model through node and graph classification tasks, using a variety of publicly available benchmark datasets. 
The proposed models are derived by incorporating three distinct smoothing techniques in the proposed models: SGCL-T (Taubin smoothing), SGCL-B (Bilateral smoothing), and SGCL-D (Diffusion-based smoothing).

We train '2-layer' GCN encoders using the SGCL framework for $200$ iterations with the Adam optimizer (learning rate $1e-3$). In the downstream task, we perform node and graph classification with $l_2$-regularized logistic regression, reporting accuracy and standard deviation after 5000 runs. For mini-batch generation, we employ the random-walk approach with a batch size of 2000, a random walk length of 4, and 3 starting root nodes. Appendix \ref{benchmark_statistic} provides comprehensive details of the experiments.
A comprehensive computational complexity analysis is also provided in Appendix \ref{Computational_analysis}.





\subsection{Node classification}
\vspace{-5pt}

In the first experiment, we evaluate the SGCL model on six small to medium-scale benchmark datasets: Cora, Citeseer, Pubmed, CoauthorCS, Computers, and Photo. Table \ref{tab:table_smal_mini_batch} presents the performance results in comparison with baseline models. To generate mini-batches in this experiment, we utilize a random-walk sampling, as outlined in Appendix \ref{sampling_alg}. A summary of the results derived from other mini-batching approaches is reported in Table \ref{tab:mini_batches_approaches}.

The results indicate that the SGCL model outperforms the state-of-the-art on most benchmarks, validating the effectiveness of our learning framework. On the "Computers" graph, which has a notably high average node degree but lower homophily (Table \ref{tab:table_datasets_node}), the influence of neighboring nodes in the smoothing approaches is reduced, leading to performance degradation compared to CGRA.


\begin{table*}[ht]
\caption{Comparison of node classification accuracies of proposed models vs. baselines on small and medium-scaled graphs (mean ± std).}
\label{tab:table_smal_mini_batch}
\begin{center}
\begin{small}
\resizebox{0.9\linewidth}{!}{%
\begin{tabular}{lcccccc} 
\toprule
\textbf{Model} & \textbf{Cora} & \textbf{Citeseer} & \textbf{Pubmed} & \textbf{CoauthorCS} & \textbf{Computers} & \textbf{Photo} \\ 
\midrule

DGI  \cite{veličković2018deep} & 82.3±0.6 & 71.8±0.7 & 76.8±0.6 & 92.15±0.63 & 83.95±0.47 & 91.61±0.22 \\
GRACE \cite{grace2020vf} & 83.3±0.4 & 72.1±0.5 & 73.63±0.20 & 91.12±0.20 & 89.53±0.35 & 92.78±0.45 \\
MVGRL \cite{icml2020_1971} & 83.11±0.12 & 73.3±0.5 & 84.27±0.04 & 92.11±0.12 & 87.52±0.11 & 91.74±0.07 \\
BGRL \cite{thakoor2021bootstrapped} & 83.77± 0.57 & 73.07±0.06 & 84.62±0.35 & 93.31±0.13 & \underline{90.34±0.19} & 93.17±0.3 \\
G-BT \cite{BIELAK2022109631} & 83.63±0.44 & 72.95±0.17 & 84.52±0.12 & 92.95±0.17 & 88.14±0.33 & 92.63±0.44 \\
CGRA \cite{DUAN2023223} & 83.8±0.4 & 69.23±1.19 & 82.8±0.4 & 92.8±0.5 & \textbf{90.5±0.4} & 92.4±0.2 \\
GRLC \cite{GRLC2023} & 83.5±0.5 & 72.6±0.6 & 82.1±0.4 & 90.36±0.27 & 88.54±0.23 & 92.3±0.5 \\
ProGCL-weight \cite{progcl2023} & 81.91±0.12 & 69.24±0.21 & \underline{84.89±0.04} & 93.51±0.06 & 89.28±0.15 & 93.30±0.09 \\
ProGCL-mix \cite{progcl2023} & 83.71±0.04 & 68.38±0.3 & 84.64±0.03 & \underline{93.67±0.12} & 89.55±0.16 & \underline{93.64±0.13} \\
GraphMAE2 \cite{graphMAE-hou2023} &  \underline{84.5±.0.6} & 73.4±0.3 & 81.4±0.5 & -- & -- & -- \\
AUGCL \cite{10382709} & -- & -- & -- & -- & 88.94±0.44 & 93.43±0.32 \\
GREET \cite{liu2023beyond} & 83.81±0.87 & 73.08±0.84 & 80.29±1.00 & \textbf{94.65±0.18} & 87.94±0.35 & 92.85±0.31  \\

\rowcolor{gray!20}SGCL-T & 84.33±0.45 & \underline{74.94±0.79} & 84.25±0.35 & 92.25±0.15 & 87.21±0.42 & 93.12±0.7 \\
\rowcolor{gray!20} SGCL-B & \textbf{84.78±0.3} & 74.30±1.4 & 84.1±0.25 & 92.33±0.4 & 89.75±0.8 & \textbf{93.72±0.12} \\
\rowcolor{gray!20} SGCL-D & 84.17±0.43 & \textbf{75.72±0.59} & \textbf{85.12±0.3} & 92.14±0.26 & 86.11±0.3 & 92.87±0.6 \\
\bottomrule
\end{tabular}
}
\end{small}
\end{center}
\end{table*}

The observed performance verifies the enhanced capacity achieved through the utilization of the geometric structure inherent in graphs, enabling improved exploration of positive and negative pairs within the conventional contrastive learning framework. \mb{More evaluation on heterophilic graphs can be found in Appendix \ref{heterophilic_evaluation}.}

Furthermore, we evaluate the proposed framework on three large-scale graphs: ogbn-arxiv, ogbn-products, and ogbn-proteins. Here, the importance of the mini-batch generation step becomes more prominent, as full-batch processing of large-scale graphs can impose considerable demands on GPU memory by requiring all node embeddings to be loaded onto the GPU. We employ a random-walk sampling approach to generate mini-batches.

\vspace{-5pt}
\begin{wraptable}{r}{0.5\textwidth}
\caption{Comparison of node classification accuracies of proposed models vs. baselines on large-scaled graphs (mean ± std).}
\label{tab:table_large_mini_batch}
\begin{center}
\begin{small}
\resizebox{1\linewidth}{!}{%
\begin{tabular}{lccc}
\toprule
\textbf{Model} & \textbf{ogbn-arxiv} & \textbf{ogbn-products} & \textbf{ogbn-proteins}\\ 
\midrule
DGI  \cite{veličković2018deep} & 67.07±0.5 & 68.68±0.6 & \underline{94.11±0.1} \\ 
GRACE \cite{grace2020vf} & 67.92±0.4 & 72.10±0.7 & 94.11±0.2 \\ 
MVGRL \cite{icml2020_1971} & 60.68±0.5 & 69.90±0.9 & 93.87±0.3 \\
BGRL \cite{thakoor2021bootstrapped} & 63.88±0.2 & 66.23±0.5 & 92.94±0.3 \\ 
GBT \cite{BIELAK2022109631} & 69.05±0.3 & 65.74±0.4 & 94.07±0.3 \\ 
GraphMAE2 \cite{graphMAE-hou2023} &  68.95±0.4 & 74.32±0.5 & -- \\

\rowcolor{gray!20} SGCL-T & \textbf{70.89±0.2} & \textbf{75.97±0.1} & \textbf{94.64±0.2} \\
\rowcolor{gray!20} SGCL-B & \underline{70.34±0.4} & \underline{74.33±0.4} & 93.55±0.2 \\
\rowcolor{gray!20} SGCL-D & 70.52±0.3 & 74.15±0.2 & 93.19±0.1 \\ 
\bottomrule
\end{tabular}
}
\end{small}
\end{center}
\end{wraptable}

The results presented in Table \ref{tab:table_large_mini_batch}, demonstrate that the SGCL consistently outperforms other contrastive learning methods on large-scale graphs. Rssults for GraphMAE2 have been reproduced using the standard data split to ensure fair comparisons.
It's worth noting that ogbn-products serves as a valuable benchmark for our proposed models providing two key advantages. Firstly, its high homophily rate increases the likelihood of identifying neighboring nodes of positive pairs as new positive pairs, thereby enhancing the performance of the model. Secondly, by using mini-batch graphs instead of the full-batch graph with numerous connected components, we can effectively bypass the extremely small components. This approach offers richer neighboring information, leading to the generation of more effective augmented graphs and enhancing the performance of the contrastive loss framework.

\subsection{Graph classification}
\label{secLgraph_classification}
\vspace{-5pt}
Graph classification is another important downstream task, employed to reflect the effectiveness of the learned graph representation. In this experiment, we follow the InfoGraph \cite{sun2019infograph} setting for graph classification and compare the accuracy with self-supervised state-of-the-art methods. The results reported in Table \ref{tab:table_graph_classification} indicate that, in comparison to the best-performing state-of-the-art methods, the proposed model demonstrates enhanced accuracy for IMDB-BINARY, PROTEINS, and ENZYMES, while maintaining comparable accuracy on other benchmarks. It's worth mentioning that the accuracies of all models are reported from their respective published papers, except for the BGRL results, which we reproduced under the same experimental setting.

\begin{table*}[h]
\centering
\caption{Comparison of graph classification accuracies of proposed models vs. baselines.}
\label{tab:table_graph_classification}
\begin{center}
\begin{small}
\resizebox{0.7\linewidth}{!}{%
\begin{tabular}{lccccc}
\toprule
\textbf{Model} & \textbf{IMDB-Binary} & \textbf{PTC-MR} & \textbf{MUTAG} & \textbf{PROTEINS} & \textbf{ENZYMES} \\ \midrule
InfoGraph \cite{sun2019infograph} & 73.0±0.9 & 61.7±1.4 & 89.0±1.1 & 74.4±0.3 & 50.2±1.4 \\
GraphCL \cite{NEURIPS2020_3fe23034} & 71.1±0.4 & 63.6±1.8 & 86.8±1.3 & 74.4±0.5 & 55.1±1.6 \\
MVGRL \cite{icml2020_1971} & 74.2±0.7 & 62.5±1.7 & 89.7±1.1 & 71.5±0.3 & 48.3±1.2 \\
AD-GCL \cite{suresh2021adversarial} & 71.5±1.0 & 61.2±1.4 & 86.8±1.3 & 75.0±0.5 & 42.6±1.1 \\
BGRL \cite{thakoor2021bootstrapped} & 72.8±0.5 & 57.4±0.9 & 86.0±1.8 & 77.4±2.4 & 50.7±9.0 \\
LaGraph \cite{pmlr-v162-xie22e} & 73.7±0.9 & 60.8±1.1 & \underline{90.2±1.1} & 75.2±0.4 & 40.9±1.7 \\
ProGCL-mix \cite{progcl2023} & 71.6±0.6 & -- & 88.7±1.4 & 74.5±0.4 & -- \\
CGRA \cite{DUAN2023223} & \underline{75.6±0.5} & \textbf{65.7±1.8} & \textbf{91.1±2.5} & 76.2±0.6 & 61.1±0.9 \\ 
AUGCL \cite{10382709} & 72.4±0.8 & -- & 89.2±1.0 & 75.7±0.4 & --\\

\rowcolor{gray!20} SGCL-T & 75.2±2.8 & \underline{64.0±1.6} & 89.0±2.3 & 79.4±1.9 & \textbf{65.3±3.6} \\
\rowcolor{gray!20} SGCL-B & 73.2±3.7 & 62.5±1.8 & 87.0±2.8 & \textbf{81.6±2.3} & 63.7±1.6 \\
\rowcolor{gray!20} SGCL-D & \textbf{75.8±1.9} & 62.6±1.4 & 86.0±2.6 & \underline{81.5±2.3} & \underline{64.3±2.2} \\ 
\bottomrule
\end{tabular}%
}
\end{small}
\end{center}
\end{table*}

\vspace{-10pt}
\section{Conclusion}
Conventional Graph Contrastive Learning (GCL) methods use a straightforward approach for distinguishing positive and negative pairs, often leading to challenges in uniformly identifying negative pairs regardless of their proximity. In this paper, we introduced a Smooth Graph Contrastive Learning (SGCL) method, which incorporates the geometric structure of graph views into a smoothed contrastive loss function. SGCL offers an intuitive way that employs three smoothing approaches to consider proximity information when assigning positive and negative pairs. 
The GCL framework is enhanced for large-scale graphs by incorporating a mini-batch strategy, leading to improved model efficiency and computational scalability. 
The evaluations, conducted on graphs of varying scales, consistently show that SGCL outperforms state-of-the-art GCL approaches in node and graph classification tasks. This emphasizes the effectiveness of the smoothed contrastive loss function in capturing and utilizing proximity information, ultimately improving the performance of the SGCL.

\section*{Acknowledgements}

The authors acknowledge the anonymous reviewers for their valuable suggestions and Johannes Lutzeyer for insightful discussions. Parts of this work were supported by the ERC Consolidator Grant 101087347 (VEGA) and the ANR AI Chair AIGRETTE.



\bibliographystyle{unsrtnat}
\bibliography{reference}

\begin{thebibliography}{57}
\providecommand{\natexlab}[1]{#1}
\providecommand{\url}[1]{\texttt{#1}}
\expandafter\ifx\csname urlstyle\endcsname\relax
  \providecommand{\doi}[1]{doi: #1}\else
  \providecommand{\doi}{doi: \begingroup \urlstyle{rm}\Url}\fi

\bibitem[Gilmer et~al.(2017)Gilmer, Schoenholz, Riley, Vinyals, and Dahl]{pmlr-v70-gilmer17a}
Justin Gilmer, Samuel~S. Schoenholz, Patrick~F. Riley, Oriol Vinyals, and George~E. Dahl.
\newblock Neural message passing for quantum chemistry.
\newblock In Doina Precup and Yee~Whye Teh, editors, \emph{Proceedings of the 34th International Conference on Machine Learning}, volume~70 of \emph{Proceedings of Machine Learning Research}, pages 1263--1272. PMLR, 06--11 Aug 2017.
\newblock URL \url{https://proceedings.mlr.press/v70/gilmer17a.html}.

\bibitem[Kipf and Welling(2017)]{kipf2017semisupervised}
Thomas~N. Kipf and Max Welling.
\newblock Semi-supervised classification with graph convolutional networks.
\newblock In \emph{International Conference on Learning Representations}, 2017.
\newblock URL \url{https://openreview.net/forum?id=SJU4ayYgl}.

\bibitem[Xu et~al.(2019{\natexlab{a}})Xu, Hu, Leskovec, and Jegelka]{xu2018how}
Keyulu Xu, Weihua Hu, Jure Leskovec, and Stefanie Jegelka.
\newblock How powerful are graph neural networks?
\newblock In \emph{International Conference on Learning Representations}, 2019{\natexlab{a}}.
\newblock URL \url{https://openreview.net/forum?id=ryGs6iA5Km}.

\bibitem[Veličković et~al.(2018)Veličković, Cucurull, Casanova, Romero, Liò, and Bengio]{veličković2018graph}
Petar Veličković, Guillem Cucurull, Arantxa Casanova, Adriana Romero, Pietro Liò, and Yoshua Bengio.
\newblock Graph attention networks.
\newblock In \emph{International Conference on Learning Representations}, 2018.
\newblock URL \url{https://openreview.net/forum?id=rJXMpikCZ}.

\bibitem[Behmanesh et~al.(2023)Behmanesh, Krahn, and Ovsjanikov]{pmlr-v202-behmanesh23a}
Maysam Behmanesh, Maximilian Krahn, and Maks Ovsjanikov.
\newblock {TIDE}: Time derivative diffusion for deep learning on graphs.
\newblock In Andreas Krause, Emma Brunskill, Kyunghyun Cho, Barbara Engelhardt, Sivan Sabato, and Jonathan Scarlett, editors, \emph{Proceedings of the 40th International Conference on Machine Learning}, volume 202 of \emph{Proceedings of Machine Learning Research}, pages 2015--2030. PMLR, 23--29 Jul 2023.
\newblock URL \url{https://proceedings.mlr.press/v202/behmanesh23a.html}.

\bibitem[Liu et~al.(2023{\natexlab{a}})Liu, Zhang, Hou, Mian, Wang, Zhang, and Tang]{9462394}
Xiao Liu, Fanjin Zhang, Zhenyu Hou, Li~Mian, Zhaoyu Wang, Jing Zhang, and Jie Tang.
\newblock Self-supervised learning: Generative or contrastive.
\newblock \emph{IEEE Transactions on Knowledge and Data Engineering}, 35\penalty0 (1):\penalty0 857--876, 2023{\natexlab{a}}.
\newblock \doi{10.1109/TKDE.2021.3090866}.

\bibitem[Wu et~al.(2023)Wu, Lin, Tan, Gao, and Li]{9632431}
Lirong Wu, Haitao Lin, Cheng Tan, Zhangyang Gao, and Stan~Z. Li.
\newblock Self-supervised learning on graphs: Contrastive, generative, or predictive.
\newblock \emph{IEEE Transactions on Knowledge and Data Engineering}, 35\penalty0 (4):\penalty0 4216--4235, 2023.
\newblock \doi{10.1109/TKDE.2021.3131584}.

\bibitem[Xie et~al.(2023)Xie, Xu, Zhang, Wang, and Ji]{9764632}
Yaochen Xie, Zhao Xu, Jingtun Zhang, Zhengyang Wang, and Shuiwang Ji.
\newblock Self-supervised learning of graph neural networks: A unified review.
\newblock \emph{IEEE Transactions on Pattern Analysis and Machine Intelligence}, 45\penalty0 (2):\penalty0 2412--2429, 2023.
\newblock \doi{10.1109/TPAMI.2022.3170559}.

\bibitem[Li et~al.(2019)Li, Gu, Dullien, Vinyals, and Kohli]{pmlr-v97-li19d}
Yujia Li, Chenjie Gu, Thomas Dullien, Oriol Vinyals, and Pushmeet Kohli.
\newblock Graph matching networks for learning the similarity of graph structured objects.
\newblock In Kamalika Chaudhuri and Ruslan Salakhutdinov, editors, \emph{Proceedings of the 36th International Conference on Machine Learning}, volume~97 of \emph{Proceedings of Machine Learning Research}, pages 3835--3845. PMLR, 09--15 Jun 2019.
\newblock URL \url{https://proceedings.mlr.press/v97/li19d.html}.

\bibitem[Veli{\v{c}}kovi{\'{c}} et~al.(2019)Veli{\v{c}}kovi{\'{c}}, Fedus, Hamilton, Li{\`{o}}, Bengio, and Hjelm]{veličković2018deep}
Petar Veli{\v{c}}kovi{\'{c}}, William Fedus, William~L. Hamilton, Pietro Li{\`{o}}, Yoshua Bengio, and R~Devon Hjelm.
\newblock {Deep Graph Infomax}.
\newblock In \emph{International Conference on Learning Representations}, 2019.
\newblock URL \url{https://openreview.net/forum?id=rklz9iAcKQ}.

\bibitem[Zhu et~al.(2020)Zhu, Xu, Yu, Liu, Wu, and Wang]{grace2020vf}
Yanqiao Zhu, Yichen Xu, Feng Yu, Qiang Liu, Shu Wu, and Liang Wang.
\newblock {Deep Graph Contrastive Representation Learning}.
\newblock In \emph{ICML Workshop on Graph Representation Learning and Beyond}, 2020.
\newblock URL \url{http://arxiv.org/abs/2006.04131}.

\bibitem[Hassani and Khasahmadi(2020)]{icml2020_1971}
Kaveh Hassani and Amir~Hosein Khasahmadi.
\newblock Contrastive multi-view representation learning on graphs.
\newblock In \emph{Proceedings of International Conference on Machine Learning}, pages 3451--3461, 2020.

\bibitem[Xia et~al.(2021)Xia, Wu, Wang, Chen, and Li]{Xia2021ProGCLRH}
Jun Xia, Lirong Wu, Ge~Wang, Jintao Chen, and Stan~Z. Li.
\newblock Progcl: Rethinking hard negative mining in graph contrastive learning.
\newblock In \emph{International Conference on Machine Learning}, 2021.
\newblock URL \url{https://api.semanticscholar.org/CorpusID:249282506}.

\bibitem[van~den Oord et~al.(2019)van~den Oord, Li, and Vinyals]{Oord2018}
Aaron van~den Oord, Yazhe Li, and Oriol Vinyals.
\newblock Representation learning with contrastive predictive coding, 2019.
\newblock URL \url{https://arxiv.org/abs/1807.03748}.

\bibitem[Taubin(1995)]{Taubin}
Gabriel Taubin.
\newblock A signal processing approach to fair surface design.
\newblock In \emph{Proceedings of the 22nd Annual Conference on Computer Graphics and Interactive Techniques}, SIGGRAPH '95, page 351–358, New York, NY, USA, 1995. Association for Computing Machinery.
\newblock ISBN 0897917014.
\newblock \doi{10.1145/218380.218473}.
\newblock URL \url{https://doi.org/10.1145/218380.218473}.

\bibitem[Tomasi and Manduchi(1998)]{710815}
C.~Tomasi and R.~Manduchi.
\newblock Bilateral filtering for gray and color images.
\newblock In \emph{Proceedings of the Sixth International Conference on Computer Vision}, ICCV '98, page 839, USA, 1998. IEEE Computer Society.
\newblock ISBN 8173192219.

\bibitem[Gerig et~al.(1992)Gerig, Kubler, Kikinis, and Jolesz]{PALMA20143887}
G.~Gerig, O.~Kubler, R.~Kikinis, and F.A. Jolesz.
\newblock Nonlinear anisotropic filtering of mri data.
\newblock \emph{IEEE Transactions on Medical Imaging}, 11\penalty0 (2):\penalty0 221--232, 1992.
\newblock \doi{10.1109/42.141646}.

\bibitem[Shuman et~al.(2013)Shuman, Narang, Frossard, Ortega, and Vandergheynst]{6494675}
David~I Shuman, Sunil~K. Narang, Pascal Frossard, Antonio Ortega, and Pierre Vandergheynst.
\newblock The emerging field of signal processing on graphs: Extending high-dimensional data analysis to networks and other irregular domains.
\newblock \emph{IEEE Signal Processing Magazine}, 30\penalty0 (3):\penalty0 83--98, 2013.
\newblock \doi{10.1109/MSP.2012.2235192}.

\bibitem[Karimi et~al.(2018)Karimi, Génois, Wagner, Singer, and Strohmaier]{Karimi2018}
Fariba Karimi, Mathieu Génois, Claudia Wagner, Philipp Singer, and Markus Strohmaier.
\newblock Homophily influences ranking of minorities in social networks.
\newblock \emph{Scientific Reports}, 8:\penalty0 1--12, 2018.
\newblock ISSN 2045-2322.
\newblock \doi{https://doi.org/10.1038/s41598-018-29405-7}.

\bibitem[Zeng et~al.(2020)Zeng, Zhou, Srivastava, Kannan, and Prasanna]{ZengZSKP20}
Hanqing Zeng, Hongkuan Zhou, Ajitesh Srivastava, Rajgopal Kannan, and Viktor~K. Prasanna.
\newblock Graphsaint: Graph sampling based inductive learning method.
\newblock In \emph{8th International Conference on Learning Representations, {ICLR} 2020, Addis Ababa, Ethiopia, April 26-30, 2020}. OpenReview.net, 2020.
\newblock URL \url{https://openreview.net/forum?id=BJe8pkHFwS}.

\bibitem[Thakoor et~al.(2022)Thakoor, Tallec, Azar, Azabou, Dyer, Munos, Veli{\v{c}}kovi{\'c}, and Valko]{thakoor2021bootstrapped}
Shantanu Thakoor, Corentin Tallec, Mohammad~Gheshlaghi Azar, Mehdi Azabou, Eva~L Dyer, Remi Munos, Petar Veli{\v{c}}kovi{\'c}, and Michal Valko.
\newblock Large-scale representation learning on graphs via bootstrapping.
\newblock In \emph{International Conference on Learning Representations}, 2022.
\newblock URL \url{https://openreview.net/forum?id=0UXT6PpRpW}.

\bibitem[Bielak et~al.(2022)Bielak, Kajdanowicz, and Chawla]{BIELAK2022109631}
Piotr Bielak, Tomasz Kajdanowicz, and Nitesh~V. Chawla.
\newblock Graph barlow twins: A self-supervised representation learning framework for graphs.
\newblock \emph{Knowledge-Based Systems}, 256:\penalty0 109631, 2022.
\newblock ISSN 0950-7051.
\newblock \doi{https://doi.org/10.1016/j.knosys.2022.109631}.
\newblock URL \url{https://www.sciencedirect.com/science/article/pii/S095070512200822X}.

\bibitem[Duan et~al.(2023)Duan, Xie, Li, and Tang]{DUAN2023223}
Haoran Duan, Cheng Xie, Bin Li, and Peng Tang.
\newblock Self-supervised contrastive graph representation with node and graph augmentation.
\newblock \emph{Neural Networks}, 167:\penalty0 223--232, 2023.
\newblock ISSN 0893-6080.
\newblock \doi{https://doi.org/10.1016/j.neunet.2023.08.039}.
\newblock URL \url{https://www.sciencedirect.com/science/article/pii/S0893608023004598}.

\bibitem[Peng et~al.(2023)Peng, Mo, Xu, Shen, Shi, Li, Shen, and Zhu]{GRLC2023}
Liang Peng, Yujie Mo, Jie Xu, Jialie Shen, Xiaoshuang Shi, Xiaoxiao Li, Heng~Tao Shen, and Xiaofeng Zhu.
\newblock Grlc: Graph representation learning with constraints.
\newblock \emph{IEEE Transactions on Neural Networks and Learning Systems}, pages 1--14, 2023.
\newblock \doi{10.1109/TNNLS.2022.3230979}.

\bibitem[Xia et~al.(2022)Xia, Wu, Wang, and Li]{progcl2023}
Jun Xia, Lirong Wu, Ge~Wang, and Stan~Z. Li.
\newblock Progcl: Rethinking hard negative mining in graph contrastive learning.
\newblock In \emph{International conference on machine learning}. PMLR, 2022.

\bibitem[Hou et~al.(2023)Hou, He, Cen, Liu, Dong, Kharlamov, and Tang]{graphMAE-hou2023}
Zhenyu Hou, Yufei He, Yukuo Cen, Xiao Liu, Yuxiao Dong, Evgeny Kharlamov, and Jie Tang.
\newblock Graphmae2: A decoding-enhanced masked self-supervised graph learner.
\newblock In \emph{Proceedings of the ACM Web Conference 2023}, WWW '23, page 737–746, New York, NY, USA, 2023. Association for Computing Machinery.
\newblock ISBN 9781450394161.
\newblock \doi{10.1145/3543507.3583379}.
\newblock URL \url{https://doi.org/10.1145/3543507.3583379}.

\bibitem[Niu et~al.(2024)Niu, Pang, and Chen]{10382709}
Chaoxi Niu, Guansong Pang, and Ling Chen.
\newblock Affinity uncertainty-based hard negative mining in graph contrastive learning.
\newblock \emph{IEEE Transactions on Neural Networks and Learning Systems}, pages 1--11, 2024.
\newblock \doi{10.1109/TNNLS.2023.3339770}.

\bibitem[Liu et~al.(2023{\natexlab{b}})Liu, Zheng, Zhang, Lee, and Pan]{liu2023beyond}
Yixin Liu, Yizhen Zheng, Daokun Zhang, Vincent Lee, and Shirui Pan.
\newblock Beyond smoothing: Unsupervised graph representation learning with edge heterophily discriminating.
\newblock In \emph{AAAI}, 2023{\natexlab{b}}.

\bibitem[Sun et~al.(2019)Sun, Hoffman, Verma, and Tang]{sun2019infograph}
Fan-Yun Sun, Jordan Hoffman, Vikas Verma, and Jian Tang.
\newblock Infograph: Unsupervised and semi-supervised graph-level representation learning via mutual information maximization.
\newblock In \emph{International Conference on Learning Representations}, 2019.

\bibitem[You et~al.(2020)You, Chen, Sui, Chen, Wang, and Shen]{NEURIPS2020_3fe23034}
Yuning You, Tianlong Chen, Yongduo Sui, Ting Chen, Zhangyang Wang, and Yang Shen.
\newblock Graph contrastive learning with augmentations.
\newblock In H.~Larochelle, M.~Ranzato, R.~Hadsell, M.F. Balcan, and H.~Lin, editors, \emph{Advances in Neural Information Processing Systems}, volume~33, pages 5812--5823. Curran Associates, Inc., 2020.
\newblock URL \url{https://proceedings.neurips.cc/paper_files/paper/2020/file/3fe230348e9a12c13120749e3f9fa4cd-Paper.pdf}.

\bibitem[Suresh et~al.(2021)Suresh, Li, Hao, and Neville]{suresh2021adversarial}
Susheel Suresh, Pan Li, Cong Hao, and Jennifer Neville.
\newblock Adversarial graph augmentation to improve graph contrastive learning.
\newblock In A.~Beygelzimer, Y.~Dauphin, P.~Liang, and J.~Wortman Vaughan, editors, \emph{Advances in Neural Information Processing Systems}, 2021.
\newblock URL \url{https://openreview.net/forum?id=ioyq7NsR1KJ}.

\bibitem[Xie et~al.(2022)Xie, Xu, and Ji]{pmlr-v162-xie22e}
Yaochen Xie, Zhao Xu, and Shuiwang Ji.
\newblock Self-supervised representation learning via latent graph prediction.
\newblock In Kamalika Chaudhuri, Stefanie Jegelka, Le~Song, Csaba Szepesvari, Gang Niu, and Sivan Sabato, editors, \emph{Proceedings of the 39th International Conference on Machine Learning}, volume 162 of \emph{Proceedings of Machine Learning Research}, pages 24460--24477. PMLR, 17--23 Jul 2022.
\newblock URL \url{https://proceedings.mlr.press/v162/xie22e.html}.

\bibitem[Wu et~al.(2021)Wu, Pan, Chen, Long, Zhang, and Yu]{9046288}
Zonghan Wu, Shirui Pan, Fengwen Chen, Guodong Long, Chengqi Zhang, and Philip~S. Yu.
\newblock A comprehensive survey on graph neural networks.
\newblock \emph{IEEE Transactions on Neural Networks and Learning Systems}, 32\penalty0 (1):\penalty0 4--24, 2021.
\newblock \doi{10.1109/TNNLS.2020.2978386}.

\bibitem[Bronstein et~al.(2017)Bronstein, Bruna, LeCun, Szlam, and Vandergheynst]{7974879}
Michael~M. Bronstein, Joan Bruna, Yann LeCun, Arthur Szlam, and Pierre Vandergheynst.
\newblock Geometric deep learning: Going beyond euclidean data.
\newblock \emph{IEEE Signal Processing Magazine}, 34\penalty0 (4):\penalty0 18--42, 2017.
\newblock \doi{10.1109/MSP.2017.2693418}.

\bibitem[Bianchi et~al.(2021)Bianchi, Grattarola, Livi, and Alippi]{bianchi2021graph}
Filippo~Maria Bianchi, Daniele Grattarola, Lorenzo Livi, and Cesare Alippi.
\newblock Graph neural networks with convolutional arma filters.
\newblock \emph{IEEE Transactions on Pattern Analysis and Machine Intelligence}, 2021.

\bibitem[Xu et~al.(2019{\natexlab{b}})Xu, Shen, Cao, Qiu, and Cheng]{xu2018graph}
Bingbing Xu, Huawei Shen, Qi~Cao, Yunqi Qiu, and Xueqi Cheng.
\newblock Graph wavelet neural network.
\newblock In \emph{International Conference on Learning Representations}, 2019{\natexlab{b}}.

\bibitem[Behmanesh et~al.(2022)Behmanesh, Adibi, Ehsani, and Chanussot]{behmane2021}
Maysam Behmanesh, Peyman Adibi, Sayyed Mohammad~Saeed Ehsani, and Jocelyn Chanussot.
\newblock Geometric multimodal deep learning with multiscaled graph wavelet convolutional network.
\newblock \emph{IEEE Transactions on Neural Networks and Learning Systems}, pages 1--15, 2022.
\newblock \doi{10.1109/TNNLS.2022.3213589}.

\bibitem[Chamberlain et~al.(2021)Chamberlain, Rowbottom, Gorinova, Bronstein, Webb, and Rossi]{chamberlain2021grand}
Ben Chamberlain, James Rowbottom, Maria~I Gorinova, Michael Bronstein, Stefan Webb, and Emanuele Rossi.
\newblock {GRAND:} graph neural diffusion.
\newblock In \emph{International Conference on Machine Learning}, pages 1407--1418. PMLR, 2021.

\bibitem[Perozzi et~al.(2014)Perozzi, Al-Rfou, and Skiena]{perozzi2014deepwalk}
Bryan Perozzi, Rami Al-Rfou, and Steven Skiena.
\newblock Deepwalk: Online learning of social representations.
\newblock In \emph{Proceedings of the 20th ACM SIGKDD International Conference on Knowledge Discovery and Data Mining}, 2014.

\bibitem[Grover and Leskovec(2016)]{grover2016node2vec}
Aditya Grover and Jure Leskovec.
\newblock node2vec: Scalable feature learning for networks.
\newblock In \emph{Proceedings of the 22nd ACM SIGKDD International Conference on Knowledge Discovery and Data Mining}, 2016.

\bibitem[Tang et~al.(2015)Tang, Qu, Wang, Zhang, Yan, and Mei]{tang2015line}
Jian Tang, Meng Qu, Mingzhe Wang, Ming Zhang, Jun Yan, and Qiaozhu Mei.
\newblock Line: Large-scale information network embedding.
\newblock In \emph{Proceedings of the 24th International Conference on World Wide Web}, 2015.

\bibitem[Hamilton et~al.(2017)Hamilton, Ying, and Leskovec]{hamilton2017inductive}
William~L. Hamilton, Rex Ying, and Jure Leskovec.
\newblock Inductive representation learning on large graphs.
\newblock In \emph{Advances in Neural Information Processing Systems}, 2017.

\bibitem[Ou et~al.(2016)Ou, Cui, Pei, Zhang, and Zhu]{ou2016asymmetric}
Mingdong Ou, Peng Cui, Jian Pei, Ziwei Zhang, and Wenwu Zhu.
\newblock Asymmetric transitivity preserving graph embedding.
\newblock In \emph{Proceedings of the 22nd ACM SIGKDD international conference on Knowledge discovery and data mining}, pages 1105--1114. ACM, 2016.

\bibitem[Zeng and Xie(2021)]{Zeng_Xie_2021}
Jiaqi Zeng and Pengtao Xie.
\newblock Contrastive self-supervised learning for graph classification.
\newblock \emph{Proceedings of the AAAI Conference on Artificial Intelligence}, 35\penalty0 (12):\penalty0 10824--10832, May 2021.
\newblock \doi{10.1609/aaai.v35i12.17293}.
\newblock URL \url{https://ojs.aaai.org/index.php/AAAI/article/view/17293}.

\bibitem[He et~al.(2020)He, Fan, Wu, Xie, and Girshick]{9157636}
Kaiming He, Haoqi Fan, Yuxin Wu, Saining Xie, and Ross Girshick.
\newblock Momentum contrast for unsupervised visual representation learning.
\newblock In \emph{2020 IEEE/CVF Conference on Computer Vision and Pattern Recognition (CVPR)}, pages 9726--9735, 2020.
\newblock \doi{10.1109/CVPR42600.2020.00975}.

\bibitem[Grill et~al.(2020)Grill, Strub, Altch\'{e}, Tallec, Richemond, Buchatskaya, Doersch, Pires, Guo, Azar, Piot, Kavukcuoglu, Munos, and Valko]{10.5555}
Jean-Bastien Grill, Florian Strub, Florent Altch\'{e}, Corentin Tallec, Pierre~H. Richemond, Elena Buchatskaya, Carl Doersch, Bernardo~Avila Pires, Zhaohan~Daniel Guo, Mohammad~Gheshlaghi Azar, Bilal Piot, Koray Kavukcuoglu, R\'{e}mi Munos, and Michal Valko.
\newblock Bootstrap your own latent a new approach to self-supervised learning.
\newblock In \emph{Proceedings of the 34th International Conference on Neural Information Processing Systems}, NIPS'20, Red Hook, NY, USA, 2020. Curran Associates Inc.
\newblock ISBN 9781713829546.

\bibitem[Ning et~al.(2022)Ning, Wang, Wang, Qiao, Fan, Zhang, Du, and Zhou]{Ning2022GraphSL}
Zhiyuan Ning, P.~Wang, Pengyang Wang, Ziyue Qiao, Wei Fan, Denghui Zhang, Yi~Du, and Yuanchun Zhou.
\newblock Graph soft-contrastive learning via neighborhood ranking.
\newblock \emph{ArXiv}, abs/2209.13964, 2022.
\newblock URL \url{https://api.semanticscholar.org/CorpusID:252568115}.

\bibitem[Zhu et~al.(2021{\natexlab{a}})Zhu, Yang, Xu, Wang, Zhang, and Han]{zhu2021transfer}
Qi~Zhu, Carl Yang, Yidan Xu, Haonan Wang, Chao Zhang, and Jiawei Han.
\newblock Transfer learning of graph neural networks with ego-graph information maximization.
\newblock In A.~Beygelzimer, Y.~Dauphin, P.~Liang, and J.~Wortman Vaughan, editors, \emph{Advances in Neural Information Processing Systems}, 2021{\natexlab{a}}.
\newblock URL \url{https://openreview.net/forum?id=CzVPfeqPOBu}.

\bibitem[Sen et~al.(2008)Sen, Namata, Bilgic, Getoor, Galligher, and Eliassi-Rad]{Sen_Namata_2008}
Prithviraj Sen, Galileo Namata, Mustafa Bilgic, Lise Getoor, Brian Galligher, and Tina Eliassi-Rad.
\newblock Collective classification in network data.
\newblock \emph{AI Magazine}, 29\penalty0 (3):\penalty0 93, Sep. 2008.
\newblock \doi{10.1609/aimag.v29i3.2157}.
\newblock URL \url{https://ojs.aaai.org/aimagazine/index.php/aimagazine/article/view/2157}.

\bibitem[Sinha et~al.(2015)Sinha, Shen, Song, Ma, Eide, Hsu, and Wang]{Sinha742839}
Arnab Sinha, Zhihong Shen, Yang Song, Hao Ma, Darrin Eide, Bo-June~(Paul) Hsu, and Kuansan Wang.
\newblock An overview of microsoft academic service (mas) and applications.
\newblock In \emph{Proceedings of the 24th International Conference on World Wide Web}, WWW '15 Companion, page 243–246, New York, NY, USA, 2015. Association for Computing Machinery.
\newblock ISBN 9781450334730.
\newblock \doi{10.1145/2740908.2742839}.
\newblock URL \url{https://doi.org/10.1145/2740908.2742839}.

\bibitem[McAuley et~al.(2015)McAuley, Targett, Shi, and van~den Hengel]{McAuley767755}
Julian McAuley, Christopher Targett, Qinfeng Shi, and Anton van~den Hengel.
\newblock Image-based recommendations on styles and substitutes.
\newblock In \emph{Proceedings of the 38th International ACM SIGIR Conference on Research and Development in Information Retrieval}, SIGIR '15, page 43–52, New York, NY, USA, 2015. Association for Computing Machinery.
\newblock ISBN 9781450336215.
\newblock \doi{10.1145/2766462.2767755}.
\newblock URL \url{https://doi.org/10.1145/2766462.2767755}.

\bibitem[Hu et~al.(2020)Hu, Fey, Zitnik, Dong, Ren, Liu, Catasta, and Leskovec]{osti_10396194}
Weihua Hu, Matthias Fey, Marinka Zitnik, Yuxiao Dong, Hongyu Ren, Bowen Liu, Michele Catasta, and Jure Leskovec.
\newblock Open graph benchmark: Datasets for machine learning on graphs.
\newblock In \emph{Proceedings of the 34th International Conference on Neural Information Processing Systems}, NIPS'20, Red Hook, NY, USA, 2020. Curran Associates Inc.
\newblock ISBN 9781713829546.

\bibitem[Kriege and Mutzel(2012)]{3042614}
Nils Kriege and Petra Mutzel.
\newblock Subgraph matching kernels for attributed graphs.
\newblock In \emph{Proceedings of the 29th International Coference on International Conference on Machine Learning}, ICML'12, page 291–298, Madison, WI, USA, 2012. Omnipress.
\newblock ISBN 9781450312851.

\bibitem[Yanardag and Vishwanathan(2015)]{2783258417}
Pinar Yanardag and S.V.N. Vishwanathan.
\newblock Deep graph kernels.
\newblock In \emph{Proceedings of the 21th ACM SIGKDD International Conference on Knowledge Discovery and Data Mining}, KDD '15, page 1365–1374, New York, NY, USA, 2015. Association for Computing Machinery.
\newblock ISBN 9781450336642.
\newblock \doi{10.1145/2783258.2783417}.
\newblock URL \url{https://doi.org/10.1145/2783258.2783417}.

\bibitem[Wale et~al.(2008)Wale, Watson, and Karypis]{s1011501035}
Nikil Wale, Ian~A. Watson, and George Karypis.
\newblock Comparison of descriptor spaces for chemical compound retrieval and classification.
\newblock \emph{Knowl. Inf. Syst.}, 14\penalty0 (3):\penalty0 347–375, mar 2008.
\newblock ISSN 0219-1377.
\newblock \doi{10.1007/s10115-007-0103-5}.
\newblock URL \url{https://doi.org/10.1007/s10115-007-0103-5}.

\bibitem[Borgwardt et~al.(2005)Borgwardt, Ong, Schönauer, Vishwanathan, Smola, and Kriegel]{ti1007}
Karsten~M. Borgwardt, Cheng~Soon Ong, Stefan Schönauer, S.~V.~N. Vishwanathan, Alex~J. Smola, and Hans-Peter Kriegel.
\newblock {Protein function prediction via graph kernels}.
\newblock In \emph{Bioinformatics}, volume~21, pages i47--i56, 06 2005.
\newblock \doi{10.1093/bioinformatics/bti1007}.
\newblock URL \url{https://doi.org/10.1093/bioinformatics/bti1007}.

\bibitem[Zhu et~al.(2021{\natexlab{b}})Zhu, Xu, Liu, and Wu]{Zhu:2021tu}
Yanqiao Zhu, Yichen Xu, Qiang Liu, and Shu Wu.
\newblock {An Empirical Study of Graph Contrastive Learning}.
\newblock \emph{arXiv.org}, September 2021{\natexlab{b}}.

\end{thebibliography}

\appendix

\section{Related work}
\label{app_related_work}

\subsection{Graph representation learning}
In recent years, graph neural networks (GNNs) have made significant progress, by the emergence of a multitude of methods dedicated to enhancing graph representation learning. These methods have been designed to address various aspects of network embeddings, including proximity, structure, attributes, learning paradigms, and scalability \cite{9046288, 7974879}. Among the notable GNN approaches, Graph Convolutional Networks (GCN) \cite{kipf2017semisupervised} is one of the foundational GNNs that uses convolutional operations to capture local and global information from neighboring nodes, making them effective for tasks like node classification. To overcome the constraints associated with conventional graph convolutions and their approximations, the Graph Attention Network (GAT) \cite{veličković2018graph} introduces the notion of masked self-attentional layers, thereby enhancing its capacity to capture crucial node relationships. By integrating an autoregressive moving average (ARMA) filter, GNN-ARMA \cite{bianchi2021graph} extends the functionality of GNNs to adeptly capture global graph structures. GWCN, as proposed in \cite{xu2018graph,behmane2021}, utilizes graph wavelets as spectral bases for convolution. This innovative approach enables the modeling of both local and global structural patterns within graphs. GRAND \cite{chamberlain2021grand} presents an interesting perspective on graph convolution networks (GCNs) by interpreting them as a solution to the heat diffusion equation. TIDE \cite{pmlr-v202-behmanesh23a} introduces an innovative approach to tackle the oversmoothing challenge in the message-passing-based approaches by leveraging the diffusion equation to enable efficient and accurate long-distance communication between nodes in a graph.

However, it's essential to emphasize that the majority of these methods depend on supervised data, and this can be a significant limitation in real-world applications due to the difficulties associated with acquiring labeled datasets.
Several traditional unsupervised graph representation learning methods are designed to learn meaningful representations of nodes in a graph without the need for labeled data or explicit supervision. DeepWalk \cite{perozzi2014deepwalk} employs random walks and skip-gram modeling to capture local graph structure, while node2vec \cite{grover2016node2vec} extends this approach with a versatile biased random walk strategy encompassing breadth-first and depth-first exploration. LINE\cite{tang2015line} focuses on preserving both first-order and second-order proximity information in large-scale networks, and GraphSAGE \cite{hamilton2017inductive} combines random walk sampling and aggregation to capture both local and global graph structure. HOPE \cite{ou2016asymmetric} leverages higher-order proximity information to capture structural patterns beyond pairwise node relationships in graphs.

\subsection{Graph contrastive learning}

Self-supervised learning (SSL) has emerged as a powerful paradigm for mitigating the challenges posed by expensive, limited, and imbalanced labels. It enables deep learning models to train on unlabeled data, reducing the reliance on annotated labels \cite{9764632}. 

Contrastive Learning (CL) is a popular SSL technique known for its simplicity and strong empirical performance. Its fundamental objective is to create meaningful representations by pushing dissimilar pairs apart and pulling similar pairs closer together. Graph Contrastive Learning (GCL) extends the concept of CL to the domain of graphs. However, dealing with the irregular structure of graph data presents more complex challenges in designing strategies for constructing positive and negative samples compared to CL applied to visual or natural language data \cite{9632431}.

Numerous papers have emerged to address the challenges associated with GCL. These papers primarily focus on sharing valuable insights and practical approaches for three key elements of contrastive learning: data augmentation, pretext tasks, and contrastive objective \cite{9632431}.

Deep Graph Infomax (DGI) \cite{veličković2018deep} and InfoGraph \cite{sun2019infograph} are two fundamental contrastive learning models that train a node encoder by maximizing mutual information between the node representation and the global graph representation. DGI is designed for node representation learning, whereas InfoGraph focuses on graph-level representations.

MVGRL \cite{icml2020_1971} is one of the recent GCL approaches that accomplishes the learning of both node and graph-level representations by considering two matrices, namely adjacency and diffusion, as congruent views of a standard contrastive framework.

The fundamental of the aforementioned GCL approaches is the maximization of local-global mutual information within a framework. However, they all rely on a readout function to generate the global graph embedding which this function can be overly restrictive and may not always be achievable. Moreover, for approaches like DGI \cite{veličković2018deep}, there is no guarantee that the resulting graph embedding can effectively capture valuable information from the nodes, as it may not adequately preserve the distinctive features found in node-level embeddings.

Several GCL approaches, including GRACE \cite{grace2020vf}, GraphCL \cite{NEURIPS2020_3fe23034}, and CSSL \cite{Zeng_Xie_2021}, deviate from the conventional approach of contrasting local-global mutual information. Notably, these methods do not rely on making assumptions about the use of injective readout functions to generate the graph embedding.

The effectiveness of the GCL models depends on comparing each item with many negative points \cite{9157636}. However, relying on these negative examples is problematic, especially for graphs, where defining negative samples in a meaningful manner is particularly difficult.

Various models have explored different strategies to address the issue of negative pairs. 
For instance, BGRL applies the BYOL method \cite{10.5555} to graphs as a GCL approach that does not rely on negative pairs \cite{thakoor2021bootstrapped}. Similarly, Graph Barlow Twins (GBT) avoids the necessity for explicit negative pairs by utilizing a cross-correlation-based loss function \cite{BIELAK2022109631}.

Several studies focus on more informative negative samples, often referred to as hard negative samples. These samples closely resemble the anchor but have semantic differences.
Intuitively, negative samples with different labels from the anchor, yet embedded nearby, are highly beneficial for providing significant gradient information during training. It's preferable to choose negative pairs with very similar representations, as this makes it challenging for the current embedding to differentiate between them effectively.
ProGCL \cite{progcl2023} incorporates hard negative nodes into the contrastive loss to enhance performance. It applies a beta mixture model (BMM) to the pairwise similarities between the negatives and the anchor, estimating the probability of a negative being a true one. It subsequently integrates the estimated probability with the pairwise similarity to measure the hardness of the negative samples. 
AUGCL \cite{10382709} is another hard negative mining GCL model that uses an affinity-based uncertainty estimator to evaluate the hardness of negative nodes relative to each anchor node. It constructs a discriminative model using pairwise affinities between negative nodes and the anchor, identifying nodes with higher uncertainty as hard negatives. 
\mb{Graph Soft-Contrastive Learning (GSCL) \cite{Ning2022GraphSL} is a novel approach aimed at overcoming the limitations of conventional graph contrastive learning by eliminating the need for graph augmentations and negative sampling. Instead, it leverages neighborhood ranking to ensure that closer nodes are more similar to a given anchor node than those farther away, in line with the inherent structure of the graph. The key limitation of GSCL is that it employs a specialized loss function for preserving the similarity ranking which requires computing a dense geodesic distance matrix for the entire graph. This process becomes increasingly challenging as the number of hops grows, leading to significantly higher computational costs, especially in large-scale graphs. Therefore, the relative similarity concept in GSCL is particularly well-suited to homophilic graphs, where label consistency decreases with distance.}

In our approach, we neither treat negative pairs the same way as in GRACE nor ignore them like in BGRL \mb{or GSCL}. Instead, we make use of negative pairs within the contrastive loss, but with a unique approach, we use the geometric structure of graphs to effectively consider proximity among negative pairs in contrastive learning, rather than treating them all the same.  Integrating the proximity information to graph contrastive loss is still highly significant and to the best of our knowledge, our approach is the first work that addresses this limitation by promoting the geometric structure of data without encountering the limitations reported in hard negative mining methods.
Specifically, our method overcomes the necessity of computing probability distributions and does not rely on prior assumptions, such as the bimodal similarity distribution of negatives with respect to positives as observed in ProGCL.

\section{Comprehensive overview of smoothing approaches}
\label{ap_algorithms}

\subsection{Taubin smoothing}
\mb{Taubin smoothing is an iterative method that employs two distinct filters—positive and negative Laplacian filters—to enhance the smoothness of input data using the graph Laplacian matrix $\mathbf{L}$. 
Algorithm \ref{alg:taubin_smoothing} outlines this process step-by-step.}

\mb{The combination of positive and negative filters allows Taubin smoothing to balance effectively between smoothing and preserving features. It avoids excessive diffusion and oversmoothing by counteracting the smoothing effect with a corrective filter. Additionally, the parameters $\tau$ and $\mu$ enable fine control over the amount of smoothing and correction, making it adaptable to different graphs.}
\mb{However, it also has its limitations. Firstly, the method requires complex parameter tuning; balancing $\tau$ and $\mu$ effectively demands careful adjustment of hyperparameters, which can be challenging. Secondly, Taubin smoothing is sensitive to the underlying graph structure, heavily relying on the quality of the graph Laplacian. If the graph structure is irregular or contains noisy edges, the results may be adversely affected.}

\begin{algorithm}
\caption{Taubin Smoothing $\mathcal{S}_T(\mathbf{V}, \mathbf{L}; K, \mu, \tau)$}
\label{alg:taubin_smoothing}
\begin{algorithmic}[1]
\STATE \textbf{Input:} Binary matrix $\mathbf{V} \in \{0,1\}^{N \times D}$, symmetric normalized graph Laplacian matrix $\mathbf{L} \in \mathbb{R}^{N \times N}$, number of iterations $K$, negative Laplacian filter constant $\mu(<0)$, positive Laplacian filter constant $\tau(>0)$ and $\mu<-\tau$
\STATE \textbf{Output:} Smoothed matrix $\tilde{\mathbf{V}}$

\STATE \textbf{Initialize:} Set $\mathbf{V}^{(0)} \gets \mathbf{V}$

\STATE \textbf{Iterative Filtering:}
\FOR{$k \gets 1$ \textbf{to} $K$}
    \STATE \textbf{Negative Laplacian Filter:}
    \STATE \quad Compute intermediate matrix $\mathbf{V}_{\text{temp}}^{(k)}$:
    \STATE \quad \quad $\mathbf{V}_{\text{temp}}^{(k)} \gets (\mathbf{I} + \mu \mathbf{L}) \mathbf{V}^{(k-1)}$

    \STATE \textbf{Positive Laplacian Filter:}
    \STATE \quad Compute the updated matrix $\mathbf{V}^{(k)}$:
    \STATE \quad \quad $\mathbf{V}^{(k)} \gets (\mathbf{I} + \tau \mathbf{L}) \mathbf{V}_{\text{temp}}^{(k)}$
\ENDFOR

\STATE \textbf{for each node} $i \textbf{ in } \{1, 2, \ldots, N\}$
\STATE \quad If the original value $\mathbf{v}_i = 1$, then
\STATE \quad \quad Set $\tilde{\textbf{v}}_i \gets 1$

\STATE \textbf{Return} $\tilde{\mathbf{V}}$

\end{algorithmic}
\end{algorithm}

\subsection{Bilateral smoothing}
\mb{This approach smooths input data by integrating spatial proximity and intensity similarity. This method considers both the distance between nodes in the graph and the similarity of their values to achieve effective smoothing. The intensity similarity $d_{\text{int}}(i,j)$ quantifies how similar two nodes are based on their intensity or binary values, with higher similarity indicating closer values. On the other hand, spatial proximity $d_{\text{spa}}(i,j)$ refers to the distance between two nodes $i$ and $j$ in the graph, measured by the shortest path. Algorithm \ref{alg:bilateral_smoothing} presents a step-by-step outline of the process.}

\mb{Bilateral smoothing provides adaptive smoothing by responding to local differences in node values and spatial distances, enhancing its robustness in graphs with strong contrasts or noise.}
\mb{However, this approach can be computationally expensive on large graphs, as it requires calculating weights for each pair of nodes based on both spatial and intensity distances. Additionally, bilateral smoothing is sensitive to hyperparameters; the parameters $\sigma_{\text{spa}}$ and $\sigma_{\text{int}}$ must be carefully tuned, as improper values can lead to under- or over-smoothing.}

\begin{algorithm}
\caption{Bilateral Smoothing $\mathcal{S}_B(\mathbf{V}, \mathbf{A}; \sigma_{\text{spa}}, \sigma_{\text{int}})$}
\label{alg:bilateral_smoothing}
\begin{algorithmic}[1]
\STATE \textbf{Input:} Binary matrix $\mathbf{V} \in \{0,1\}^{N \times D}$, adjacency matrix $\mathbf{A} \in \mathbb{R}^{N \times N}$, spatial smoothing parameter $\sigma_{\text{spa}}$, intensity smoothing parameter $\sigma_{\text{int}}$
\STATE \textbf{Output:} Smoothed matrix $\tilde{\mathbf{V}}$

\STATE \textbf{for each pair of nodes} $(i, j)$ \textbf{in} $\{1, 2, \ldots, N\}$ where $j \in \mathcal{N}_k(i):$
\STATE \quad Compute weight $w(i,j)$:
\STATE \quad \quad $w(i,j) \gets \exp \left( - \frac{d_{\text{spa}}(i,j)}{2 \sigma_{\text{spa}}^2} - \frac{d_{\text{int}}(i,j)}{2 \sigma_{\text{int}}^2} \right)$

\STATE \textbf{for each node} $i \textbf{ in } \{1, 2, \ldots, N\}$
\STATE \quad Compute the smoothed value $\tilde{\textbf{v}}_i$:
\STATE \quad \quad $\tilde{\textbf{v}}_i \gets \frac{\sum_{j \in \mathcal{N}_k(i)} w(i,j) \textbf{v}_j}{\sum_{j \in \mathcal{N}_k(i)} w(i,j)}$

\STATE \textbf{for each node} $i \textbf{ in } \{1, 2, \ldots, N\}$
\STATE \quad If the original value $\mathbf{v}_i = 1$, then
\STATE \quad \quad Set $\tilde{\textbf{v}}_i \gets 1$

\STATE \textbf{Return} $\tilde{\mathbf{V}}$

\end{algorithmic}
\end{algorithm}

\subsection{Diffusion-based smoothing}
\mb{Diffusion-based smoothing simulates the diffusion process (similar to heat diffusion) to propagate information across the graph. The value of each node diffuses into its neighbors, gradually smoothing the graph over time. Algorithm \ref{alg:diffusion_smoothing} presents a step-by-step outline of the process.}

\mb{Diffusion-based smoothing is simple and efficient; the diffusion equation is relatively straightforward and computationally efficient to implement, making the method scalable for large graphs. Additionally, its iterative nature promotes smooth global effects, allowing values to propagate throughout the graph in a stable manner. However, there are notable disadvantages. One significant drawback is the potential loss of details; if not carefully controlled, the method can oversmooth the input, leading to the loss of sharp features or edges. Additionally, diffusion-based smoothing applies uniform smoothing; since it relies on averaging over neighboring nodes, it does not account for intensity similarity, which may result in the blurring of sharp changes in node values.}

\begin{algorithm}
\caption{Diffusion-based Smoothing $\mathcal{S}_D(\mathbf{V}, \mathbf{A}; K, \eta)$}
\label{alg:diffusion_smoothing}
\begin{algorithmic}[1]
\STATE \textbf{Input:} Binary matrix $\mathbf{V} \in \{0,1\}^{N \times D}$, adjacency matrix $\mathbf{A} \in \mathbb{R}^{N \times N}$, number of iterations $K$, diffusion rate $\eta$
\STATE \textbf{Output:} Smoothed matrix $\tilde{\mathbf{V}}$

\STATE Initialize $\mathbf{V}^{(0)} \gets \mathbf{V}$

\FOR{$k = 0$ \textbf{to} $K-1$}
    \STATE \textbf{for each node} $i \textbf{ in } \{1, 2, \ldots, N\}$
    \STATE \quad Compute the average value of neighboring nodes:
    \STATE \quad \quad $\bar{\textbf{v}}_i^{(k)} \gets \sum_{j \in \mathcal{N}(v_i)} \textbf{v}_j^{(k)}$
    \STATE \quad Update the value of node $i$:
    \STATE \quad \quad $\textbf{v}_i^{(k+1)} \gets \textbf{v}_i^{(k)} + \eta \bar{\textbf{v}}_i^{(k)}$
\ENDFOR

\STATE \textbf{for each node} $i \textbf{ in } \{1, 2, \ldots, N\}$
\STATE \quad If the original value $\mathbf{v}_i^{(0)} = 1$, then
\STATE \quad \quad Set $\textbf{v}_i^{(K)} \gets 1$

\STATE \textbf{Return} $\tilde{\mathbf{V}} \gets \mathbf{V}^{(K)}$

\end{algorithmic}
\end{algorithm}

\mb{In general, each smoothing approach has its unique strengths, making it appropriate for specific types of graph and smoothing requirements, allowing for tailored applications depending on the context. Taubin smoothing is ideal for graphs where connectivity is balanced and the local structure is not too irregular or noisy. Additionally, it is more efficient for large graphs as its computational complexity is lower compared to other methods in practice. Bilateral smoothing works well on graphs with heterogeneous or highly varying node values, such as those representing social networks, where sharp changes in node features are significant. Finally, Diffusion-based smoothing is best suited for large graphs with uniform or gradual changes, such as temperature distributions or geographical information, where computational efficiency is essential and the focus is on achieving smooth transitions without sharp features.}

\section{Mini-batch generating approaches}
\label{sampling_alg}

\noindent \textbf{Random node sampler}
approach randomly selects a subset of nodes from a given graph $\mathcal{G}=(\mathcal{V},\mathcal{E})$ according to a probability distribution $P(v)$, where $v$ represents individual nodes in the graph. The distribution $\displaystyle P(v)$ assigns a probability to each node, indicating the likelihood of that node being included in the sampled subset.

\noindent \textbf{Random edge sampler} approach randomly selects edges from a given graph $\mathcal{G}=(\mathcal{V},\mathcal{E})$ based on a predefined probability distribution. For each edge $e$ in the set of edges $\mathcal{E}$, an independent decision is made to determine whether it should be included in the subgraph $\mathcal{G}_s$. This decision is guided by a probability value $P(e)$ assigned to each edge. The sampler incorporates a budget parameter $m$ to constrain the expected number of sampled edges, ensuring that $\sum P(e) = m$, as described in  \cite{ZengZSKP20}.

\noindent \textbf{Random walk sampler} approach begins by randomly selecting $r$ root nodes as starting points on the entire graph $\mathcal{G}=(\mathcal{V},\mathcal{E})$. From each of these starting nodes, random walks of length $L$ are conducted to generate subgraphs \cite{ZengZSKP20}. To manage the potential issue of generating excessively large subgraphs, a batch size parameter $m$ is commonly employed, ensuring the approximate number of samples per batch.

\noindent \textbf{Ego graph sampler}  approach generates subgraphs centered around a specific "ego" node in a graph $\mathcal{G}=(\mathcal{V},\mathcal{E})$. This mini-batch generation approach provides a localized perspective on the graph by constructing a $k$-hop ego-graph centered at node $v_i$, where "$k$-hop" indicates that the subgraph includes nodes that can be reached within $k$ steps from $v_i$. Importantly, the sampler ensures that the maximum distance between $v_i$ and any other nodes within the ego-graph is limited to $k$, as expressed mathematically by $\forall v_j \in \mathcal{V}, \mid d(v_i,v_j)\mid<k$ \cite{zhu2021transfer}.

\section{Experimental setup}
\label{benchmark_statistic}

\subsection{Properties and statistics of the benchmarks}

For node classification, the benchmarks encompass a wide range of graph sizes, including smaller to medium-scaled ones such as Cora, Citeseer, Pubmed \cite{Sen_Namata_2008}, CoauthorCs \cite{Sinha742839}, Computers, and Photos \cite{McAuley767755}, as well as larger datasets like ogbn-arxiv, ogbn-products, ogbn-proteins, and all of which are sourced from the Open Graph Benchmark \cite{osti_10396194}. For graph classification, we employ MUTAG \cite{3042614}, PTC \cite{3042614}, IMDB-Binary \cite{2783258417}, PROTEINS \cite{s1011501035}, and ENZYMES \cite{ti1007} benchmarks.

The properties of different graph datasets used in the node and graph classification experiments are provided in Table \ref{tab:table_datasets_node} and \ref{tab:table_datasets_graph}, respectively. The homophily rate $h$ denotes the degree to which nodes in the graph connect with similar nodes (homophily) versus nodes with dissimilar nodes (heterophily). The diameter of large-scaled graphs is performed using Breadth-First Search (BFS) from a sample of 1,000 nodes selected at random.

\begin{table}[ht]
\caption{The statistics of the datasets for node classification evaluation}
\label{tab:table_datasets_node}
\begin{center}
\begin{small}
\resizebox{1\linewidth}{!}{%
\begin{tabular}{llllcccccc}
\toprule
\textbf{Scale} & \multicolumn{1}{l}{\textbf{Dataset}} & \multicolumn{1}{l}{\textbf{\#Nodes}} & \multicolumn{1}{l}{\textbf{\#Edges}} & \multicolumn{1}{l}{\textbf{\#Feature}} & \multicolumn{1}{l}{\textbf{\#Class}} & \multicolumn{1}{l}{\textbf{\#CC}} & \multicolumn{1}{l}{\textbf{h\%}} & \multicolumn{1}{l}{\textbf{Avg. N.D.}} & \multicolumn{1}{l}{\textbf{Diameter}} \\ 
\midrule
\multirow{2}{*}{Small} & Cora & 2,708 & 5,429 & 1,433 & 7 & 78 & 80.4 & 4.08 & 19 \\
 & Citeseer & 3,327 & 4,732 & 3,703 & 6 & 438 & 73.5 & 3.47 & 28 \\ \midrule
\multirow{4}{*}{Medium} & PubMed & 19,717 & 44,324 & 500 & 3 & 1 & 80.2 & 4.5 & 18 \\
 & CoauthorCs & 18,333 & 81,894 & 6,805 & 15 & 1 & 80 & 8.93 & 24 \\
 & Computers & 13,381 & 245,778 & 767 & 10 & 314 & 77.7 & 36.74 & 10 \\
 & Photos & 7,487 & 119,043 & 745 & 8 & 136 & 82.7 & 31.8 & 11 \\ \midrule
\multirow{3}{*}{Large} & ogbn-arxiv & 169,343 & 1,166,243 & 128 & 40 & 1 & 65.4 & 13.67 & 23 \\
 & ogbn-products & 2,449,029 & 61,859,140 & 100 & 47 & 52,658 & 80.8 & 51.54 & 27 \\
 & ogbn-proteins & 132,534 & 39,561,252 & 8 & 94 & 1 & 91 & 597 & 9 \\ 
 \bottomrule
 \multicolumn{10}{l}{\textbf{\#CC}: Number of connected components, \textbf{h\%}: Homophily rate, \textbf{Avg. N.D}: Average node degrees}
\end{tabular}
}
\end{small}
\end{center}
\end{table}

\begin{table}[ht]
\centering
\caption{The statistics of the datasets for graph classification evaluation}
\label{tab:table_datasets_graph}
\begin{center}
\begin{small}
\resizebox{0.6\linewidth}{!}{%
\begin{tabular}{lccccc}
\bottomrule\textbf{Dataset} & \multicolumn{1}{l}{\textbf{\#Graph}} & \multicolumn{1}{l}{\textbf{Avg. node}} & \multicolumn{1}{l}{\textbf{Avg. edge}} & \multicolumn{1}{l}{\textbf{\#Features}} & \multicolumn{1}{l}{\textbf{\#Class}} \\ 
\midrule
MUTAG & 188 & 17.9 & 39.6 & 7 & 2 \\
PTC-MR & 344 & 14.29 & 14.69 & 19 & 2 \\
IMDB-Binary & 1,000 & 19.8 & 193.1 & 1 & 2 \\
PROTEINS & 1,113 & 39.1 & 145.6 & 3 & 2 \\
ENZYMES & 600 & 32.63 & 124.3 & 3 & 6 \\ 
\bottomrule
\end{tabular}%
}
\end{small}
\end{center}
\end{table}

\subsection{Hyperparameters}
\label{details}

In all experiments, we follow the linear evaluation scheme outlined in \cite{veličković2018deep}. Initially, we start by training the '2-layer' GCN encoders using the proposed SGCL framework in an unsupervised manner. The training process consists of 200 iterations, and we utilize the Adam optimizer with a learning rate of $1e-3$.
Subsequently, the obtained embeddings are used to perform node or graph classification on a downstream task, employing a $l_2-$ regularized logistic regression classifier. The mean classification accuracy, along with the standard deviation, is then reported on the test nodes after conducting 5000 training runs.

In the mini-batch scenario of the node classification task, we employ a random-walk batch generation approach to create subgraphs from the input graph. We set the batch size to 2000 with a random walk length of 4 and 3 starting root nodes for all benchmark datasets. However, for the ogbn-products benchmark, we use a batch size of 500 with a random walk length of 20.

In the smoothing techniques, we set the parameters as follows: for Taubin smoothing, we set $\mu=-0.4$, $\tau=0.3$, and $K=2$; in the case of Bilateral smoothing, we employ $\sigma_{spa}=0.1$ and $\sigma_{init}=2$; and for Diffusion-based smoothing, we utilize $\eta=0.03$ and $K=2$.

To ensure a fair comparison with state-of-the-art models in both node and graph classification tasks, we adopt the widely used data split method from the Open Graph Benchmark, which is commonly employed in self-supervised learning. We also report values based on the respective papers. For benchmarks where experiments weren't performed in the relevant papers, we accurately reproduced their values using available code resources. It’s worth mentioning that the results on the Cora in MVGRL \cite{icml2020_1971} are reported across benchmarks with varying numbers of nodes and edges (refer to Table 1 in the respective paper). Therefore, the values are reproduced and reported using the standard Cora benchmark.

To implement the proposed model, we leveraged the extensive capabilities offered by the PyGCL library, as introduced in \cite{Zhu:2021tu}. For the graph augmentation, we employ the augmentor base class provided by PyGCL, which includes Edge Removing (ER) and Node Feature Masking (FM), both with a drop probability of $0.5$.
For a comprehensive comparison, we reported values based on the respective papers.

To ensure a fair comparison with state-of-the-art, we followed the publicly available data split of citation networks and replicated all experiments accordingly.
For benchmarks where experiments weren't performed in the relevant papers, we accurately reproduced their values using available code resources. Additionally, for large-scale graphs, we conducted experiments on most of the baselines using the PyGCL library since there was a lack of extensive baseline experimentation. The implementation is available at \url{https://github.com/maysambehmanesh/SGCL}.

All experiments are implemented using PyTorch 1.13.1 and PyTorch Geometric 2.2.0 and conducted on NVIDIA A100 GPUs with 40GB of memory.

\subsection{Baselines}

In our empirical study, we incorporate a variety of models for comparison. For node classification, these models encompass representative node classification models, as well as recently-introduced graph contrastive learning models, such as DGI \cite{veličković2018deep}, GRACE \cite{grace2020vf}, MVGRL \cite{icml2020_1971}, GBT \cite{BIELAK2022109631}, BGRL \cite{thakoor2021bootstrapped}, CGRA \cite{DUAN2023223}, and GRLC \cite{GRLC2023} serving as our baseline models. 
For graph classification, we employ seven state-of-the-art methods for graph contrastive learning, including InfoGraph \cite{sun2019infograph}, GraphCL \cite{NEURIPS2020_3fe23034}, MVGRL \cite{icml2020_1971}, BGRL \cite{thakoor2021bootstrapped}, AD-GCL \cite{suresh2021adversarial}, LaGraph \cite{pmlr-v162-xie22e}, and CGRA \cite{DUAN2023223}.




\section{Supplementary experiments}

\subsection{An empirical analysis of the feature space}
\label{emperical_feature_space}

\mb{In this section, we conduct the empirical analysis, introduced by Proposition \ref{proposi}, to validate the impact of the feature space on real-world graphs by calculating the disparity measure using two encoders within the proposed SGCL and conventional GCL (GRACE) frameworks. 
For a meaningful comparison of the disparity measures across all graphs, we normalized the values using Min-Max scaling, rescaling the measure to a range between 0 and 1.
The results presented in Table \ref{table_mds} indicate that the mean disparity of the graph encoders across all variants of SGCL is consistently lower than the conventional GCL approach. This suggests that the SGCL encoder produces node representations with greater similarity within the same class and increased distinction between different classes, reflecting more effective self-supervised learning.
}

\begin{table}[ht]
\centering
\caption{Comparison of mean disparity measures for learned features of SGCL and GCL on real-world graph benchmarks.}
\label{table_mds}
\resizebox{\textwidth}{!}{%
\begin{tabular}{lccccccc}
\toprule
Model & Cora & Citeseer & Pubmed & CoauthorCS & Computers & Photo & ogbn-arxiv \\ \midrule
GCL (GRACE) & 0.66±0.05 & 0.63±0.02 & 0.51±0.03 & 0.58±0.02 & 0.64±0.03 & 0.65±0.03 & 0.56±0.03 \\
\rowcolor{gray!20}  SGCL-T & 0.49±0.05 & 0.58±0.05 & 0.50±0.01 & 0.53±0.03 & 0.56±0.01 & 0.60±0.05 & 0.51±0.02 \\
\rowcolor{gray!20}  SGCL-B & 0.63±0.03 & \textbf{0.51±0.05} & 0.49±0.02 & 0.52±0.01 & \textbf{0.52±0.05} & 0.59±0.02 & 0.48±0.05 \\
\rowcolor{gray!20}  SGCL-D & \textbf{0.47±0.02} & 0.54±0.03 & \textbf{0.49±0.01} & \textbf{0.46±0.05} & 0.55±0.05 & \textbf{0.43±0.03} & \textbf{0.45±0.04} \\  \bottomrule

\end{tabular}%
}
\end{table}

\subsection{More evaluation on heterophilic graphs}
\label{heterophilic_evaluation}
\mb{To perform a more comprehensive analysis, we conduct experiments on graphs with varying homophily rates, utilizing different real-world graphs, as detailed in Table \ref{tab:hitr_graph}.}

\begin{table}[ht]
\centering
\caption{Different real-world graphs. The parameter $h[0,1]$ is the edge homophily ratio for homophily graphs$ h \rightarrow 1$  and for heterophily graphs $h\rightarrow 0$.}
\label{tab:hitr_graph}
\resizebox{0.82\linewidth}{!}{%
\begin{tabular}{lcccccc}
\toprule
Graph & \#Nodes & \#Edges & \begin{tabular}[c]{@{}c@{}}\#Features\end{tabular} & \#Classe & Class types & $h$ \\ \midrule
Chameleon & 2,277 & 36,101 & 2,325 & 5 & Wiki pages & 0.23 \\
Actor & 7,600 & 29,926 & 931 & 5 & Actors in movies & 0.22 \\
Cornell & 183 & 295 & 1,703 & 5 & Web pages & 0.3 \\
Texas & 183 & 309 & 1,703 & 5 & Web pages & 0.11 \\
Wisconsin & 251 & 499 & 1,703 & 5 & Web pages & 0.21 \\
Genius & 421,961 & 984,979 & 12 & 2 & marked act. & 0.618 \\
Twitch-gamers & 168,114 & 6,797,557 & 7 & 2 & mature content & 0.545 \\ \hline
\end{tabular}%
}
\end{table}

\mb{We evlauate the performance of SGCL on these graphs and compare the results with conventional GCL approaches like the GRACE model.
Results in Table \ref{tab:results_herto} indicate that the proposed models still outperform conventional GCL on heterophilic graphs. However, the advantage of smoothing methods in homophilic graphs becomes more pronounced. As the homophily rate increases, the number of false negatives also increases, emphasizing the critical role of SGCL in effectively contributing both positive and negative pairs to the contrastive loss.
}

\begin{table}[ht]
\centering
\caption{Comparison of the accuracy of proposed SGCL models on heterophilic graphs with GCL.}
\label{tab:results_herto}
\resizebox{\textwidth}{!}{%
\begin{tabular}{lccccccc}
\toprule
Model & Chameleon & Actor & Cornell & Texas & Wisconsin & Genius & Twitch-gamers \\ \midrule
GCL (GRACE) & 45.3±0.7 & 28.55±0.28 & 52.21±1.3 & 53.07±2.4 & 51.60±2.4 & 81.48±0.15 & 58.21±0.51 \\
\rowcolor{gray!20} SGCL-T & \textbf{46.32±1.3} & \textbf{29.29±0.47} & 54.45±0.6 & 54.56±1.8 & 52.50±2.3 & \textbf{82.58±0.33} & 59.37±0.47 \\
\rowcolor{gray!20} SGCL-B & 45.52±0.3 & 29.15±0.28 & \textbf{55.79±1.2} & 54.97±2.3 & 50.23±2.2 & 82.47±0.17 & \textbf{59.38±0.88} \\
\rowcolor{gray!20} SGCL-D & 45.91±2.5 & 28.74±0.76 & 53.12±5.4 & \textbf{55.48±2.7} & \textbf{53.62±2.1} & 81.86±0.2 & 58.60±0.48 \\ \hline
\end{tabular}%
}
\end{table}

\subsection{Computational analysis}
\label{Computational_analysis}

The computational cost of graph contrastive learning models is analyzed through two distinct components: pre-training and downstream task evaluation. The pre-training phase consists of mini-batch generation, augmentation generation, encoder computation, and computations of the smoothing strategy. In the downstream task phase, the model learns two input/output MLP layers and evaluates the model for tasks such as node classification.

For a graph $\mathcal{G} = (\mathcal{V}, \mathcal{E})$, with $N$ nodes and $E$ edges, the encoder computation using a message-passing-based GNN encoder $f_\theta$ efficiently computes embeddings with the complexity of $\mathcal{O}(N+E)$.
The computation cost of graph augmentation consists of applying the feature mask ($\mathcal{O}(N)$) and the edge removal mask ($\mathcal{O}(E)$). The overall complexity is $\mathcal{O}(N+E)$, with the edge removal mask being the dominant factor ($\mathcal{O}(E)$). Notably, this cost is lower than that of MVGRL \cite{icml2020_1971}, as it utilizes a Personalized PageRank-based graph diffusion approach for structural augmentations, which entails a complexity of $\mathcal{O}(I.E)$, where $I$ represents the number of iterations required for convergence.

The proposed model imposes additional computational overhead compared to the standard GCL. This includes the computation associated with the mini-batch strategy and integrating the smoothing strategy into the conventional contrastive loss.

The computation of mini-batch strategies can be disregarded in the overall complexity analysis, as it is performed offline during preprocessing. However, the complexities of mini-batch strategies to generate $k$ subgraphs with a batch size $m$ from a given graph are $\mathcal{O}(km+kE)$ for the Random Node Sampler, $\mathcal{O}(km + kE)$ for the Random Edge Sampler, $\mathcal{O}( k(r \times h × d + m))$ for the Random Walk Sampler with $r$ roots, walk length $h$ and average node degree $d$, and $\mathcal{O}(k(s ×d + m))$ for the $s$-hop Ego-graph sampler.

The main computational overhead is associated with the smoothing strategy. Taubin smoothing, which utilizes Laplacian matrices, has a computational complexity of $\mathcal{O}(N+E)$ constructing the Laplacian matrix and $\mathcal{O}(N)$ per iteration for matrix multiplication. The overall complexity, influenced by the number of iterations, ranges from linear to quadratic concerning the number of nodes $N$ and edges $E$.

The computational complexity of bilateral smoothing, which considers both spatial proximity and intensity similarity for each node, is predominantly influenced by the node degrees and the total number of edges, typically in the order $\mathcal{O}(N+E)$.

The computational complexity of diffusion-based smoothing primarily depends on the number of nodes $N$ and the number of iterations $K$. Each iteration involves summing the values of neighboring nodes, which can be considered $\mathcal{O}(deg(i))$ for each node $i$, where $deg(i)$ is the degree of node $i$. Therefore, the overall computational complexity can be expressed as $\mathcal{O}(K×N+E)$, where $K$ is the number of iterations, $N$ is the number of nodes, and $E$ is the total number of edges in the graph.

\begin{table}[ht]
\centering
\caption{ Runtime performance comparison of the proposed model and baselines across graphs of different scales (each value denotes the running time of individual epochs, measured in seconds).}
\label{tab_computation}
\begin{center}
\begin{small}
\resizebox{0.7\linewidth}{!}{%
\begin{tabular}{llccc}
\toprule
\textbf{Model} & \textbf{Phase} & \begin{tabular}[c]{@{}c@{}}\textbf{Small} \\ (Cora)\end{tabular} & \begin{tabular}[c]{@{}c@{}}\textbf{Medium}\\  (CoauthorCS)\end{tabular} & \begin{tabular}[c]{@{}c@{}}\textbf{Large}\\  (ogbn-arxiv)\end{tabular} \\ \midrule
\multirow{2}{*}{DGI} & pre-training & 0.0391 & 0.0916 & 0.0732 \\
 & downstream & 0.0024 & 0.0148 & 0.0837 \\ \midrule
\multirow{2}{*}{GRACE} & pre-training & 0.0713 & 0.3186 & 0.4233 \\
 & downstream & 0.0024 & 0.0148 & 0.0845 \\ \midrule
\multirow{2}{*}{MVGRL} & pre-training & 0.2266 & 0.7824 & 0.9407 \\
 & downstream & 0.0024 & 0.0148 & 0.0833 \\ \midrule
\multirow{2}{*}{BGRL} & pre-training & 0.0927 & 0.1849 & 0.1755 \\
 & downstream & 0.0024 & 0.0149 & 0.0846 \\ \midrule
\multirow{2}{*}{GBT} & pre-training & 0.0343 & 0.1387 & 0.5388 \\
 & downstream & 0.0024 & 0.0148 & 0.0844 \\ \midrule
 \multirow{2}{*}{GRLC} & pre-training & 0.1193 & 0.3249 & 0.5747 \\
 & downstream & 0.0682 & 0.2685 & 0.4325 \\ \midrule
\multirow{2}{*}{ProGCL-weight} & pre-training & 0.0929 & 0.3428 & -- \\
 & downstream & 0.0032 & 0.0152 & -- \\ \midrule
\multirow{2}{*}{ProGCL-mix} & pre-training & 0.1192 & 0.4993 & -- \\
 & downstream & 0.0029 & 0.0152 & -- \\ \midrule
\multirow{2}{*}{GREET} & pre-training & 0.1793 & 2.5811 & -- \\
 & downstream & 0.0031 & 0.0125 & -- \\ \midrule
\multirow{2}{*}{SGCL-T} & pre-training & 0.1133 & 0.9723 & 1.3921 \\
 & downstream & 0.0025 & 0.0149 & 0.0841 \\ \midrule
\multirow{2}{*}{SGCL-B} & pre-training & 0.9374 & 2.5303 & 3.0016 \\
 & downstream & 0.0025 & 0.0151 & 0.0848 \\ \midrule
\multirow{2}{*}{SGCL-D} & pre-training & 1.0073 & 2.6681 & 3.1296 \\
 & downstream & 0.0024 & 0.0151 & 0.0841 \\ 
 \bottomrule
\end{tabular}%
}
\end{small}
\end{center}
\end{table}

For numerical evaluation, we conduct the computational analysis to evaluate the runtime performances of three variants of the SGCL model,  comparing them with several baseline methods across graphs of varying scales. The results of these experiments are summarized in Table \ref{tab_computation}.
\mb{For this analysis, we included the majority of baselines for which implemented code was accessible; benchmarks with unreasonable runtimes are marked with a dash (–) in the table.}

These results indicate that during pre-training, SGCL-T on the small-scale graph outperforms MVGRL, \mb{GRLC, ProGCL-mix, and GREET} in running time. For medium-scale and large-scale graphs, the computational costs are approximately 18\% and 40\% higher than those of MVGRL, respectively. The computational cost of the other variants of the model is increased compared to the baselines. This observed computational overhead is associated with the expectations, as SGCL integrates supplementary information into the conventional contrastive loss function, and the smoothing strategies require the exploration of the graph to identify proximity information.

\mb{Notably, the run-time performances in the downstream evaluation phase across baselines implemented in the PyGCL library, which follows a framework similar to SGCL,} are nearly identical on each benchmark. This implies that, despite the more computation time in the pre-training phase, our model performs effectively in the downstream evaluation phase.

\mb{Additionally, we have conducted memory consumption comparisons for these baselines across graphs of different sizes, employing various graph batch generation methodds: Random Walk Sampler (RWS), Ego Graph Sampler (EGS), Random Node Sampler (RNS), and Random Edge Sampler (RES). The reported values reflect the GPU memory required (in MB) to train the models.}

\mb{Table \ref{tab:memory_cons} indicate that the proposed SGCL models utilizing NSP and ESP demonstrate better memory efficiency, outperforming most baselines, particularly on medium and large-scale graphs. 
The EGS strategy tends to be more memory-intensive than other sampling methods. This becomes a crucial consideration when working with small and medium-scale graphs, particularly when comparing our approach to more memory-efficient baselines such as DGI, BGRL, and GBT.}

\begin{table}[ht]
\centering
\caption{Memory consumption comparison between the proposed model and baselines on graphs of varying scales (MB).}
\label{tab:memory_cons}
\resizebox{0.5\linewidth}{!}{%
\begin{tabular}{llrrr}
\toprule
\textbf{Model} & & \begin{tabular}[c]{@{}c@{}}\textbf{Small} \\ (Cora)\end{tabular} & \begin{tabular}[c]{@{}c@{}}\textbf{Medium}\\  (CoauthorCS)\end{tabular} & \begin{tabular}[c]{@{}c@{}}\textbf{Large}\\  (ogbn-arxiv)\end{tabular} \\ \midrule
\multicolumn{2}{l}{DGI} & 711 & 3023 & 15993 \\ \midrule
\multicolumn{2}{l}{GRACE} & 1447 & 34255 & 34907 \\ \midrule
\multicolumn{2}{l}{MVGRL} & 2753 & 33641 & 35191 \\ \midrule
\multicolumn{2}{l}{BGRL} & 901 & 9931 & 26691 \\ \midrule
\multicolumn{2}{l}{GBT} & 971 & 5093 & 26943 \\ \midrule
\multicolumn{2}{l}{GRLC} & 1139 & 11533 & 35263 \\ \midrule
\multicolumn{2}{l}{ProGCL-weight} & 1113 & 14129 & -- \\ \midrule
\multicolumn{2}{l}{ProGCL-mix} & 1647 & 28239 & -- \\ \midrule
\multicolumn{2}{l}{GREET} & 1655 & 33167 & -- \\ \midrule
\multirow{4}{*}{SGCL-T} & RWS & 1957 & 11629 & 11639 \\
 & EGS & 2097 & 32782 & 38641 \\
 & NSP & 1075 & 2001 & 1489 \\
 & ESP & 1481 & 4249 & 3605 \\ \midrule
\multirow{4}{*}{SGCL-B} & RWS & 1973 & 9113 & 11293 \\
 & EGS & 2529 & 32137 & 38759 \\
 & NSP & 1075 & 1999 & 1473 \\
 & ESP & 1481 & 4321 & 4497 \\ \midrule
\multirow{4}{*}{SGCL-D} & RWS & 1973 & 11409 & 12051 \\
 & EGS & 2119 & 32871 & 35725 \\
 & NSP & 1977 & 2033 & 1477 \\
 & ESP & 1477 & 4715 & 3163 \\ \midrule
\end{tabular}%
}
\end{table}

\section{Ablation study}
\label{abblation}

\subsection{Evaluating with other mini-batching generation methods}
\label{app_results_mini_batches}

We conduct node classification experiments employing other mini-batching generation methods, including random node-sampling, random edge-sampling, and Ego-graph. A summary of the results derived from these mini-batching approaches is reported in Table \ref{tab:mini_batches_approaches}.

\begin{table}[ht]
\caption{Accuracy comparison of proposed models with various mini-batching generation approaches (mean ± std).}
\label{tab:mini_batches_approaches}
\begin{center}
\begin{small}
\resizebox{0.9\linewidth}{!}{%
\begin{tabular}{clllllll} 
\toprule
\multicolumn{1}{l}{\textbf{Model}} & \textbf{Sampling method} & \textbf{Cora} & \textbf{Citeseer} & \textbf{Pubmed} & \textbf{CoauthorCS} & \textbf{Computers} & \textbf{Photo} \\
\midrule
\multirow{4}{*}{SGCL-T} & RW-sampler & 84.33±0.4 & \textbf{74.94±0.8} & 84.25±0.3 & \textbf{92.25±0.1} & \textbf{87.21±0.4} & 93.12±0.7 \\
 & Ego-graph & 84.21±0.3 & 73.88±1.6 & \textbf{84.47±0.7} & 92.12±0.3 & 86.7±0.6 & 93.05±0.7 \\
 & Node-sampler & 84.12±0.8 & 74.12±1.3 & 84.14±0.4 & 92.17±0.7 & 87.08±0.4 & 92.84±1.2 \\
 & Edge-sampler & \textbf{84.53±.5} & 73.54±1.7 & 83.76±0.6 & 91.83±0.8 & 86.88±1.5 & \textbf{93.33±0.8} \\
\midrule
\multirow{4}{*}{SGCL-B} & RW-sampler & \textbf{84.78±0.3} & \textbf{74.30±1.4} & 84.1±0.2 & \textbf{92.33±0.4} & \textbf{89.75±0.8} & \textbf{93.72±0.1} \\
 & Ego-graph & 84.63±0.7 & 73.26±0.7 & \textbf{84.16±0.5} & 91.68±0.7 & 89.07±1.6 & 93.13±0.4 \\
 & Node-sampler & 84.39±0.8 & 73.74±1.4 & 83.84±1.1 & 92.18±0.3 & 88.25±0.8 & 92.22±0.6 \\
 & Edge-sampler & 84.11±1.6 & 74.22±1.5 & 83.79±0.5 & 92.22±0.6 & 88.84±0.2 & 92.16±1.2 \\
\midrule
\multirow{4}{*}{SGCL-D} & RW-sampler & 84.17±0.4 & \textbf{75.72±0.8} & 85.12±0.3 & 92.14±0.2 & \textbf{86.11±0.3} & 92.87±0.6 \\
 & Ego-graph & 84.15±1.5 & 73.87±0.6 & 84.73±0.2 & 92.17±0.6 & 84.24±0.5 & 92.26±0.5 \\
 & Node-sampler & 84.23±1.3 & 74.43±1.3 & 84.68±1.3 & 92.05±0.3 & 85.38±0.4 & 91.63±1.4 \\
 & Edge-sampler & \textbf{84.75±1.4} & 73.55±0.8 & \textbf{85.17±0.6} & \textbf{92.21±0.1} & 85.3±0.7 & \textbf{93.76±0.2} \\ 
 
\bottomrule
\end{tabular}
}
\end{small}
\end{center}
\end{table}

\vspace{15pt}
\subsection{Influence of different terms in contrastive loss function}
\label{abaltion_con_obj}

Since the number of non-zero values in $\tilde{\Pi}_{\text{neg}}^{(i,j)}$ exceeds those in $\tilde{\Pi}_{\text{pos}}^{(i,j)}$, we initially assign $\lambda$ as $1/2N$. 
This adjustment aims to achieve a trade-off between positive and negative pairs within the loss function \ref{equ_cont}. However, in the experiments, we determined its optimal value through grid search. For instance, on the Photo dataset, the optimal value for $\lambda$ was found to be around $2.3e-4$. This value aligns with our first initialization when considering the batch size of $N=2000$ in the experiments.

To perform an ablation study on the contrastive loss function, we evaluate the significance of each term of Equation \ref{equ_cont} and subsequently combine them with hyperparameter $\lambda$.
Table \ref{table_ablation_obj} provides the accuracies of different variants of SGCL achieved by different components of the contrastive loss function on three benchmarks of varying scales: small (Cora), medium (CoauthorCS), and large (ogbn-arxiv). 
Initially, we observe that the exclusion of any term from the loss function results in deteriorated or collapsed solutions, aligning with our expectations. Subsequently, we investigated the influence of the combination of two individual terms using an optimal value of $\lambda$.

\begin{table}[ht]
\centering
\caption{Accuracies of different SGCL variants influenced by individual components of the contrastive loss function Equation \ref{equ_cont}.}
\label{table_ablation_obj}
\begin{center}
\begin{small}
\resizebox{0.7\linewidth}{!}{%
\begin{tabular}{llcccc}
\toprule
\textbf{Model} & \textbf{Benchmark}  &  \textbf{(A)} & \textbf{(B)} & $\mathcal{L}_{\text{SGCL}}^{(i,j)}$ & $(\lambda)$ \\ 
\midrule
 & small (Cora) & 84.12±0.7 & 83.24±1.2 & 84.33±0.4 & (4e-4) \\
SGCL-T & medium (CoauthorCS) & 92.15±0.3 & 91.74±0.4 & 92.25±0.1 & (1e-4) \\
 & large (ogbn-arxiv) & 68.92±0.0 & 67.05±0.0 & 69.30±0.5 & (1e-4) \\ \midrule
 & small (Cora) & 84.27±0.8 & 83.84±0.7 & 84.78±0.3 & (4e-4) \\
SGCL-B & medium (CoauthorCS) & 91.73±0.4 & 91.66±0.7 & 92.33±0.4 & (1e-4) \\
 & large (ogbn-arxiv) & 68.73±0.3 & 68.29±0.4 & 69.24±0.3 & (1e-4) \\
 \midrule
 & small (Cora) & 83.9±1.3 & 83.82±1.4 & 84.17±0.4 & (4e-4) \\
SGCL-D & medium (CoauthorCS) & 91.63±0.4 & 91.32±0.6 & 92.14±0.2 & (1e-4) \\
 & large (ogbn-arxiv) & 68.40±0.3 & 68.29±0.3 & 69.03±0.4 & (1e-4) \\
 \midrule
\multicolumn{6}{l}{\textbf{(A)}: $\parallel {\tilde{\Pi}_{\text{pos}}^{(i,j)}\odot (1-\mathbf{C}^{(i,j)})} \parallel_F^2$ } \\
\multicolumn{6}{l}{\textbf{(B)}: $\parallel (1-\tilde{\Pi}_{\text{pos}}^{(i,j)}) \odot \mathbf{C}^{(i,j)} \parallel_F^2$ } \\ 
\bottomrule
 \end{tabular}%
}
\end{small}
\end{center}
\end{table}

\subsection{Ablation analysis of hyperparameters in smoothing approaches}
\label{ablation_smoothing}

\subsubsection{Taubin smoothing}
\mb{Taubin smoothing is a combined process that alternates between negative and positive Laplacian filters to smooth the signal. The negative Laplacian filter, with hyperparameter $\mu < 0$, smooths the input matrix $\mathbf{V}$, while the positive Laplacian filter, with hyperparameter $\tau>0$ (where $\mu < -\tau)$, prevents oversmoothing by restoring some of the original values.
The smoothing process alternates $K$ times between these two filters. The values of $\tau$, $\mu$, and $K$ are carefully chosen to balance the smoothing process with the positive correction, ensuring a stable result that avoids both excessive noise and oversmoothing.}

\mb{For the ablation study, we begin by fixing $K=2$ and evaluate the impact of varying the hyperparameters $\tau$ (where $0 < \tau < 1$) and $\mu$ (with $\mu < -\tau$).}
\mb{For a more detailed evaluation, we focus on the interaction between the parameters $\tau$ and $\mu$, both of which are varied over finer increments. Specifically, we evaluate the performance of the model on the Cora and Pubmed by adjusting $\tau$ in the range of $0.1$ to $0.5$ and $\mu$ between $-0.2$ and $-0.6$ $(\mu < - \tau)$. Table \ref{tab:ablation_taubin} highlights the results for different combinations of $\tau$ and $\mu$.}

\begin{table}[ht]
\centering
\caption{Performance of SGCL-T with fixed $K=2$ and different combinations of $\mu$ and $\tau$.}
\label{tab:ablation_taubin}
\resizebox{0.9\linewidth}{!}{%
\begin{tabular}{crccccc|ccccc}
\toprule
\multicolumn{1}{l}{} & \multicolumn{1}{l}{} & \multicolumn{5}{c|}{Cora} & \multicolumn{5}{c}{Pubmed} \\ \midrule 
\multicolumn{1}{l}{} & \multicolumn{1}{l}{$\mu$} & -0.2 & -0.3 & -0.4 & -0.5 & -0.6 & -0.2 & -0.3 & -0.4 & -0.5 & -0.6 \\ \midrule
\multirow{5}{*}{$\tau$} & 0.1 & \textbf{84.87} & 84.31 & 84.06 & 84.21 & 83.71 & 83.9 & 83.22 & 83.54 & 82.56 & 81.82 \\
 & 0.2 & - & 83.55 & 84.31 & 84.37 & 84.03 & - & 83.36 & 84.32 & 82.22 & 82.06 \\
 & 0.3 & - & - & 84.33 & 83.25 & 83.81 & - & - & 84.25 & \textbf{84.4} & 82.83 \\
 & 0.4 & - & - & - & 83.86 & 83.81 & - & - & - & 83.38 & 82.76 \\
 & 0.5 & - & - & - & - & 84.11 & - & - & - & - & 82.26 \\ \hline
\end{tabular}%
}
\end{table}

\mb{Similarly, we evaluate the performance of SGCL-T by varying the number of iteration $K$, while keeping $\tau = 0.3$ and $\mu = -0.4$ fixed. The evaluation is performed on both the Cora and Pubmed datasets, with the corresponding accuracy and runtime for different values of $K$. The results are summarized in Table \ref{tab:abblation_K}.}

\begin{table}[ht]
\centering
\caption{Performance of SGCL-T with fixed $\tau = 0.3$ and $\mu = -0.4$ and different numbers of iteration $K$.}
\label{tab:abblation_K}
\resizebox{0.84\linewidth}{!}{%
\begin{tabular}{llccccccc}
\toprule
 & K & 1 & 2 & 3 & 4 & 5 & 6 & 7 \\ \midrule
\multirow{2}{*}{Cora} & Accuracy & 84.25 & \textbf{84.47} & 83.55 & 84.37 & 84.11 & 84.26 & 84.11 \\ 
 & Time (s) & 0.0849 & 0.1242 & 0.1727 & 0.2183 & 0.2758 & 0.3211 & 0.3734 \\ \midrule
\multirow{2}{*}{Pubmed} & Accuracy & \textbf{84.34} & 84.25 & 84.1 & 83.42 & 83.35 & 83.76 & 83.76 \\
 & Time (s) & 0.6179 & 1.0517 & 1.4551 & 1.9107 & 2.2914 & 2.6974 & 3.1376 \\ \midrule
\end{tabular}%
}
\end{table}

\mb{From the ablation study, we observe that the highest accuracy across different benchmarks is achieved with varying hyperparameter values. However, for consistency, we set $\tau = 0.3$, $\mu = -0.4$, and $K=2$ for all experiments. It's also worth noting that as $K$ increases, computation time per epoch significantly rises for both datasets, highlighting a trade-off between accuracy and efficiency.}

\subsubsection{Bilateral smoothing}
\mb{In the bilateral smoothing approach, the $\sigma_\text{int}$ parameter controls how sensitive the bilateral filter is to intensity differences. In this context, it determines how sharply the filter distinguishes between different intensities. When $\sigma_\text{int}$ increases, the sensitivity of the filter to differences in intensity decreases, causing it to treat all intensity values more similarly. This makes the bilateral filter act more like a standard Gaussian filter, which smooths uniformly without considering intensity differences. Similarily, the $\sigma_\text{spa}$ parameter determines the extent to which spatial proximity affects the bilateral smoothing process. It controls how much weight is given to neighboring nodes based on their distance in the graph. When $\sigma_\text{spa}$ is small, only nodes that are very close to each other have a strong influence on each other during smoothing. This preserves fine details and small features. As $\sigma_\text{spa}$ increases, the filter begins to smooth over larger distances, meaning that larger features in the graph get smoothed out. In this case, even nodes that are farther apart will start influencing each other more, resulting in a broader, more generalized smoothing effect.}

\mb{For SGCL-B, we conduct an ablation study on two key hyperparameters: $\sigma_{\text{init}}$ and $\sigma_{\text{spa}}$. The study is carried out on two datasets, Cora and Pubmed, with varying values for both hyperparameters. The results are presented in Table \ref{tab:ablation_bilateral}, where we evaluate model performance for different combinations of $\sigma_{\text{init}}$ and $\sigma_{\text{spa}}$.}

\begin{table}[ht]
\centering
\caption{Performance of SGCL-B with different combinations of $\sigma_{\text{init}}$ and $\sigma_{\text{spa}}$.}
\label{tab:ablation_bilateral}
\resizebox{0.9\linewidth}{!}{%
\begin{tabular}{lcccccc|ccccc}
\toprule
 & \multicolumn{1}{l}{} & \multicolumn{5}{c|}{Cora} & \multicolumn{5}{c}{Pubmed} \\ \midrule
 & \multicolumn{1}{l}{$\sigma_{\text{spa}}$} & 0.01 & 0.05 & 0.1 & 0.15 & 0.2 & 0.01 & 0.05 & 0.1 & 0.15 & 0.2 \\ \midrule
\multirow{5}{*}{$\sigma_{\text{init}}$} & 1 & 84.48 & \textbf{84.72} & 84.87 & 83.96 & 84.26 & 83.08 & 84.11 & 84.25 & 83.12 & 82.14 \\
 & 2 & 84.37 & 84.16 & 84.78 & 84.16 & 83.4 & 83.25 & 84.23 & 84.1 & 82.49 & 83.96 \\
 & 3 & 84.47 & 82.99 & 84.21 & 84.21 & 84.47 & 84.22 & \textbf{84.26} & 84.17 & 83.26 & 83.28 \\
 & 4 & 84.47 & 84.37 & 83.96 & 83.91 & 83.15 & 83.86 & 83.92 & 84.09 & 83.56 & 83.87 \\
 & 5 & 84.26 & 83.86 & 85.28 & 84.01 & 83.35 & 83.63 & 83.89 & 83.82 & 83.24 & 83.8 \\ \midrule
\end{tabular}%
}
\end{table}

\mb{These results indicate that the choice of hyperparameters impacts the performance of the model, and optimal settings vary across different datasets. Additionally, smaller values of $\sigma_{\text{spa}}$ tend to preserve local graph features better, while larger values induce more smoothing, which affects accuracy differently across datasets.}

\subsubsection{Diffusion-based smoothing}

\mb{In this approach, the diffusion rate parameter $\eta$ controls the speed of smoothing. Larger values lead to faster diffusion and more aggressive smoothing, while smaller values retain more of the original structure. The parameter $K$ controls how long the smoothing process continues. More iterations lead to a more globally smooth result, while fewer iterations keep the smoothing more localized.
Essentially, diffusion-based smoothing gradually spreads information from each node to its neighbors, helping to equalize values across the graph. The choice of $\eta$ and $K$ allows for fine control over how quickly and broadly this smoothing occurs, making it a flexible approach for graph-based data processing.}

\mb{In the ablation study for SGCL-D, we examine the impact of two key hyperparameters: $\eta$ (diffusion rate) and $K$  (number of iterations) on the Cora and Pubmed datasets. The results are summarized in Table \ref{tab:ablation_diffuaion}.}

\vspace{-10pt}
\begin{table}[ht]
\centering
\caption{Performance of SGCL-D with different combinations of $\eta$ and $K$.}
\label{tab:ablation_diffuaion}
\resizebox{0.9\linewidth}{!}{%
\begin{tabular}{lcccccc|ccccc}
\toprule
 & \multicolumn{1}{l}{} & \multicolumn{5}{c|}{Cora} & \multicolumn{5}{c}{Pubmed} \\ \midrule
 & \multicolumn{1}{l}{$\eta$} & 0.01 & 0.03 & 0.1 & 0.2 & 0.5 & 0.01 & 0.03 & 0.1 & 0.2 & 0.5 \\ \midrule
\multirow{5}{*}{$K$} & 1 & 84.26 & 84.67 & 84.12 & 84.22 & 84.11 & 84.23 & 84.29 & 84.17 & 83.17 & 83.46 \\
 & 2 & 84.52 & 84.47 & 84.53 & 84.11 & 84.23 & 84.11 & 85.12 & 83.65 & 84.12 & 83.25 \\
 & 3 & 84.33 & 84.17 & 84.23 & 83.9 & 83.87 & 84.63 & \textbf{85.17} & 84.27 & 83.63 & 84.04 \\
 & 4 & 84.85 & 84.52 & 83.86 & 83.27 & 83.66 & 84.55 & 84.29 & 83.66 & 83.26 & 83.28 \\
 & 5 & \textbf{84.97} & 84.28 & 83.54 & 83.64 & 83.78 & 84.57 & 83.88 & 83.46 & 83.7 & 83.68 \\ \hline
\end{tabular}%
}
\end{table}

\mb{On Cora, the highest accuracy $(84.97$\%) is achieved with $\eta = 0.01$ and $K=5$, indicating that slower diffusion and more iterations tend to yield better results. For Pubmed, the best accuracy ($85.17$\%) is achieved with $\eta = 0.03$ and $K=3$, showing that moderate diffusion rates paired with a balanced number of iterations deliver optimal results.
These results highlight a trade-off between diffusion speed and the number of iterations: smaller $\eta$ values combined with more iterations generally retain more local details, while larger $\eta$ values spread smoothing effects faster but may reduce accuracy due to over-smoothing.}

\section{Limitations and future directions}

\mb{As discussed, contrasting with true hard negatives—nodes with different labels from the anchor but located nearby—can improve model performance by providing strong gradient signals and enhancing contrast. However, in many benchmarks, particularly in homophilic graphs, these nearby nodes are likely to share the same label as the anchor, making them potential false negatives. This issue can undesirably push away semantically similar samples, resulting in performance degradation.
The proposed SGCL methods effectively mitigate the harmful impact of false negative nodes. However, this also inevitably reduces the influence of potential true hard negative nodes.
Despite this trade-off, the overall reduction in false negatives is expected to have a more significant positive effect than the reduced influence of true hard negatives, as supported by our analytical analysis.}

\mb{As the first to introduce smoothed positive/negative pairs for graph contrastive learning, we explored the development of a stable and effective learnable smoothing objective. However, we found that a straightforward learnable solution is challenging to train since a learnable smoothing objective can make the overall loss unstable. Consequently, the basic version did not yield performance improvements. Nonetheless, we recognize this as a crucial direction for future enhancement and view it as an exciting avenue for further research.}

\end{document}